\documentclass{article}
\usepackage{amssymb}
\usepackage{multicol}
\usepackage[bookmarks=true]{hyperref}
\usepackage{graphicx}
\usepackage{pgfplots} 
\usepackage{pgf-pie}
\usepackage[dvipsnames]{xcolor}
\usepackage{wrapfig}   
\usepackage{caption}
\usepackage{algpseudocode}
\usepackage{amsmath}
\usepackage{amssymb}
\usepackage{mathtools}
\usepackage{placeins}
\usepackage{subcaption}
\usepackage{lipsum}   
\usepackage{algorithm}
\usepackage{algpseudocode}
\usepackage[hypcap=true]{caption}
\usepackage{authblk}
\usepackage{multirow}
\usepackage{multicol}
\usepackage{booktabs}  

\usepackage{float}
\usepackage{cuted}
\usepackage{capt-of}

\makeatletter
\renewcommand\AB@affilsepx{\quad \protect\Affilfont}
\makeatother

\usepackage{subcaption}

\usetikzlibrary{calc} 
\usepackage[final]{corl_2025} 

\usepackage{xparse}
\usepackage[normalem]{ulem}

\usepackage{amssymb}

\usepackage{array}
\usepackage{multirow}
\usepackage{booktabs}
\usepackage{makecell}

\NewDocumentCommand\NewIndexedVar{mmm}{
\expandafter\NewDocumentCommand\csname#1\endcsname{se{^_}}{#2\IfNoValueTF{##2}{}{^{##2}}\IfNoValueTF{##3}{\IfBooleanTF{##1}{}{_#3}}{_{##3}}}
}

\NewDocumentCommand\itime{}{n}

\NewDocumentCommand\goal{}{\boldsymbol{g}}
\NewDocumentCommand\emb{}{\boldsymbol{e}}

\NewDocumentCommand\p{mm}{#1\left(#2\right)}

\NewDocumentCommand\act{s o}{\boldsymbol{\IfBooleanTF{#1}{a}{\bar{a}}}\IfNoValueTF{#2}{}{_{#2}}}
\NewDocumentCommand\state{s o}{\boldsymbol{\IfBooleanTF{#1}{s}{\bar{s}}}\IfNoValueTF{#2}{}{_{#2}}}

\usepackage[acronym]{glossaries}
\newacronym{flore}{FLOWER}{\textbf{Flo}rence \textbf{W}ith \textbf{E}mbodied Flow}
\newacronym{vlm}{VLM}{Vision-Language-Model}
\newacronym{mlp}{MLP}{Multi-Layer Perceptron}
\newacronym{llm}{LLM}{Large-Language-Model}
\newacronym{vla}{VLA}{Vision-Language-Action-Model}
\newacronym{vae}{VAE}{Variational Autoencoder}
\newacronym{mse}{MSE}{Mean Squared Error}
\newacronym{vit}{ViT}{Vision Transformer}
\newacronym{davit}{DaViT}{Dual-Attention Vision Transformer}
\newacronym{oxe}{OXE}{Open-X-Embodiment}
\newacronym{modit}{MoRiT}{Mixture-of-Rectified Flow Transformer}
\newacronym{il}{IL}{Imitation Learning}

\let\oldgls\gls
\let\oldglspl\glspl

\renewcommand{\gls}[1]{\oldgls*{#1}}
\renewcommand{\glspl}[1]{\oldglspl*{#1}}

\title{FLOWER: Democratizing Generalist Robot Policies with Efficient Vision-Language-Action Flow Policies}

\author[1]{Moritz Reuss}
\author[1]{Hongyi Zhou}
\author[1]{Marcel Rühle}
\author[1]{Ömer Erdinç Yağmurlu}
\author[2]{\authorcr Fabian Otto}
\author[1]{Rudolf Lioutikov} 
\affil[1]{Intuitive Robots Lab, Karlsruhe Institute of Technology, Germany}
\affil[2]{Microsoft Research}

\begin{document}
\maketitle

\newcommand\blfootnote[1]{%
  \begingroup
  \renewcommand\thefootnote{}\footnote{#1}%
  \addtocounter{footnote}{-1}%
  \endgroup
}
\renewcommand{\subsectionautorefname}{Subsection}
\blfootnote{Correspondence to: moritz.reuss@kit.edu}


\begin{abstract}
Developing efficient Vision-Language-Action (VLA) policies is crucial for practical robotics deployment, yet current approaches face prohibitive computational costs and resource requirements. Existing diffusion-based VLA policies require multi-billion-parameter models and massive datasets to achieve strong performance. We tackle this efficiency challenge with two contributions: intermediate-modality fusion, which reallocates capacity to the diffusion head by pruning up to $50\%$ of LLM layers, and action-specific Global-AdaLN conditioning, which cuts parameters by $20\%$ through modular adaptation. We integrate these advances into a novel 950 M-parameter VLA called FLOWER. Pretrained in just 200 H100 GPU hours, FLOWER delivers competitive performance with bigger VLAs across $190$ tasks spanning ten simulation and real-world benchmarks and demonstrates robustness across diverse robotic embodiments. In addition, FLOWER achieves a new SoTA of 4.53 on the CALVIN ABC benchmark. Demos, code and pretrained weights are available at \url{https://intuitive-robots.github.io/flower_vla/}.
\end{abstract}
\keywords{Imitation Learning, VLA,Language-conditioned Manipulation}

\section{Introduction}

\begin{figure*}
    \centering
    \begin{subfigure}{\linewidth}
        \centering
        \includegraphics[width=1.0\linewidth]{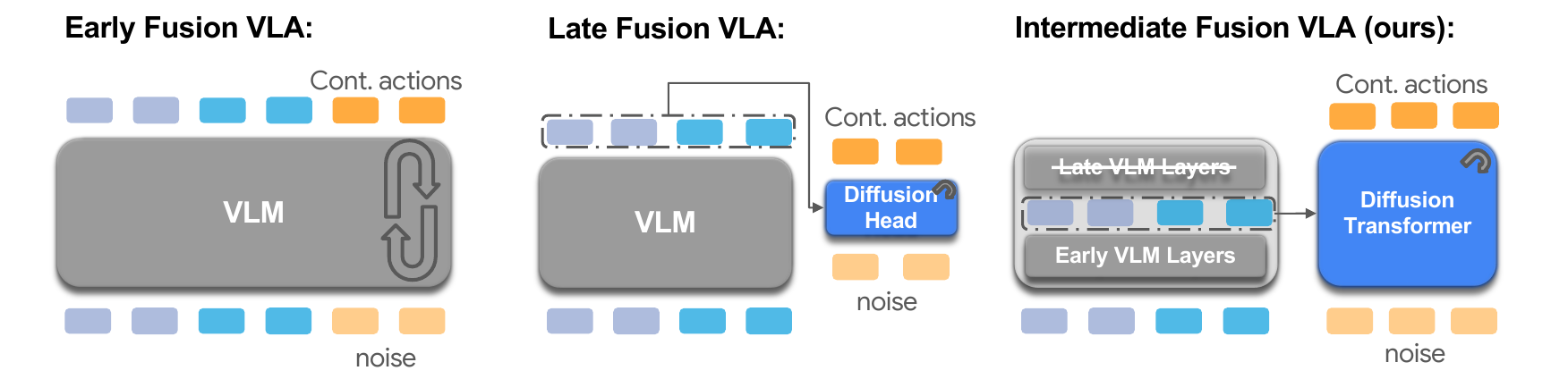}
        \label{fig:fusion_strategies}
    \end{subfigure}
    
    \vspace{-1.5em} 
    \begin{subfigure}[b]{0.32\textwidth}
        \centering
        \includegraphics[width=\textwidth]{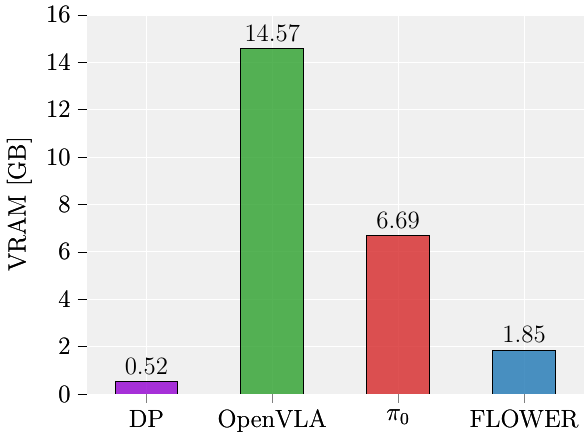}
        \caption{\textbf{Small GPU Memory}}
        \label{fig:small_memory}
    \end{subfigure}
    \hfill 
    \begin{subfigure}[b]{0.32\textwidth}
        \centering
        \includegraphics[width=\textwidth]{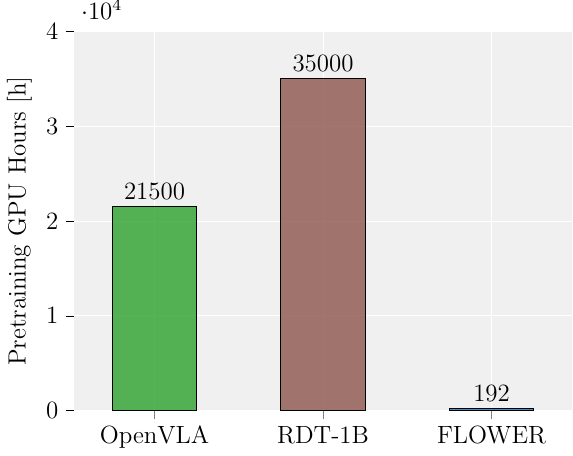}
        \caption{\textbf{Less Compute}}
        \label{fig:less_compute}
    \end{subfigure}
    \hfill 
    \begin{subfigure}[b]{0.32\textwidth}
        \centering
        \includegraphics[width=\textwidth]{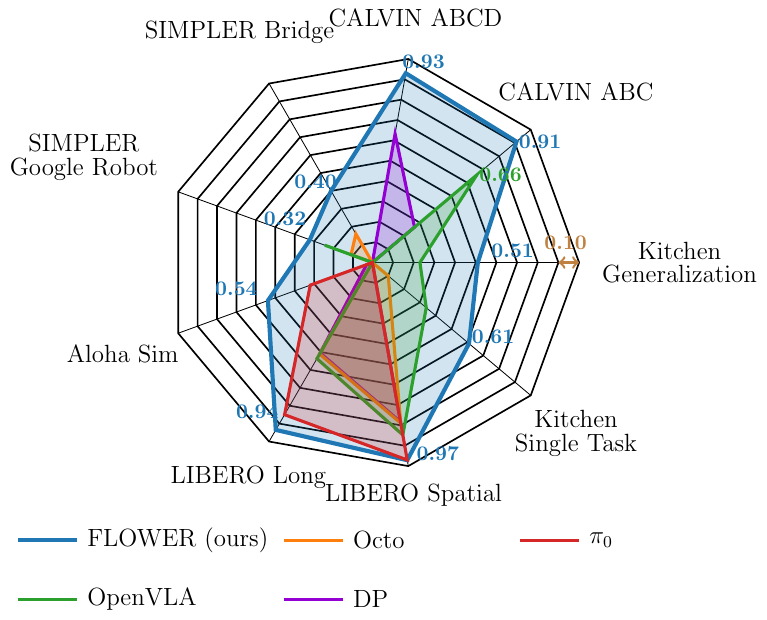}
        \caption{\textbf{Strong Performance}}
        \label{fig:strong_performance}
    \end{subfigure}
    \caption{\textbf{Intermediate fusion for efficient VLA policies.} Our fusion strategy (top-right) strategically prunes VLM layers while enhancing Flow Transformer capacity in parameter-constrained settings. This approach informs FLOWER, a novel, $950M$ VLA that achieves competitive performance across 10 benchmarks using only $1\%$ of the pretraining compute of models like OpenVLA~\cite{kim2024openvla}, while maintaining a small memory footprint across diverse embodiments and action spaces (bottom).}
    \label{fig:combined_figure1}
\end{figure*}

Generalist robotic manipulation policies that execute diverse tasks across different embodiments remain a key goal in robotics. 
Recent advances in \gls{il} have made significant progress toward this vision, particularly along the direction of generalist \gls{vla} Policies \cite{rt12022arxiv, kim2024openvla, octo_2023}.
\glspl{vla} fine‑tune pretrained \glspl{vlm} to generate robot actions from free-form language commands \cite{rt12022arxiv,kim2024openvla,octo_2023}. 
These models commonly adopt discrete \cite{li2023generalist,kim2024openvla,driess2023palm,open_x_embodiment_rt_x_2023} or diffusion‑based objectives \cite{black2024pi_0, liu2024rdt, bjorck2025gr00t} for action prediction. 
In particular, diffusion‑ and flow‑based action generation excel at modeling complex, multimodal action distributions and have been successfully adopted for \glspl{vla} \cite{liu2024rdt,black2024pi_0,reuss2024efficient}.
While \glspl{vla} offer many advantages, they have a major limitation: Current \glspl{vla} such as OpenVLA \cite{kim2024openvla} and RDT‑1B \cite{liu2024rdt} contain several billion parameters and they require a lot of compute for pretraining, fine‑tuning and real‑robot deployment.  
This barrier limits access to more diverse research in the field.

In this work, we present contributions and insights that result in an efficient \gls{vla} policy that contains fewer than 1 billion parameters while matching current SoTA \glspl{vla} on $190$ tasks across $10$ benchmarks with $4$ embodiments. 
As illustrated in Figure~\ref{fig:combined_figure1}, dedicating most parameters to a deep, off‐the‐shelf VLM backbone forces the diffusion head to be severely under‐parameterized for modeling rich, multimodal robot trajectories; conversely, shrinking the \gls{vlm} to free up capacity strips away semantic features essential for robust instruction conditioning. Moreover, relying on the full \gls{vlm} for denoising significantly slows convergence during training and increases inference latency.

To resolve this budget–tradeoff, we introduce four key contributions. First, we propose \emph{intermediate‐level fusion}: we prune between $30\%$ and $50\%$ of the pretrained \gls{vlm}’s layers and condition our Flow Transformer on the resulting intermediate embeddings, thereby retaining semantic grounding while reclaiming parameters. 
Second, we develop a \emph{global action‐space AdaLN} mechanism: an action‐specific LayerNorm controller within the diffusion transformer that reduces head parameters by 20\% without any loss of accuracy or expressivity.  
Third, we provide extensive ablations across multiple benchmarks that evaluate design choices and yield additional insights into which \gls{vlm} architectures and pretraining objectives are best suited for efficient \gls{vla} adoption.

Leveraging these contributions and insights, we present our fourth contribution: \textbf{\gls{flore}}, a $950$M parameter \gls{vla} policy that converges quickly, cuts pretraining cost on heterogeneous robotics data by $99\%$, and yields competitive performance with current \gls{vla}s across $190$ tasks in $10$ benchmarks across simulation and real world settings.


\section{Related Work}

Early imitation-learning studies demonstrated that policy performance scales with dataset size. For example, \citet{pinto2016supersizing} used over 50K grasps and observed steady improvements in manipulation proficiency.
Building on this, the OXE benchmark provided 1.4M trajectories across more than 20 robot embodiments, enabling research into generalist policies \cite{open_x_embodiment_rt_x_2023}.
Several diffusion-policy approaches directly train on OXE without leveraging pretrained \glspl{vlm}. 
Octo \cite{octo_2023} applies a transformer-based diffusion policy to delta end-effector actions but lacks a pretrained vision–language encoder and model capacity, which limits its generalization as observed on benchmarks such as SIMPLER \cite{li24simpler}. RDT introduces a 1.2B-parameter diffusion transformer paired with an 11.4B-parameter pretrained language encoder model and \gls{vit}. However, its one-month pretraining on 48 A100 GPUs highlights the need for more computationally efficient \gls{vla} pretraining \cite{liu2024rdt}. Addressing this need, our contributions yield versatile \glspl{vla} in only $200$ GPU hours of pretraining. 

To address generalization, recent methods incorporate large pretrained \gls{vlm}s into the policy. OpenVLA fine-tunes a 7.7B-parameter \gls{vlm} for discrete end-effector action prediction, yet its size makes real-robot deployment challenging \cite{kim2024openvla}. 
Several other \gls{vla}s apply discrete action prediction \cite{rt12022arxiv, brohan2023rt, driess2023palm, bousmalis2023robocat, shafiullah2023bringing, etukuru2024robot}.
RoboDual improves upon OpenVLA, by integrating a small diffusion transformer with OpenVLA via asynchronous updates, while Latent Bridge uses a fine-tuned \gls{vlm} to generate latent commands that condition a diffusion policy \cite{bu2024towards, shentu2024llms}. 
Similar to FLOWER, 
$\pi_0$ \cite{black2024pi_0} and GR00T-N1 \cite{bjorck2025gr00t}, also introduce generalist flow-based VLA with more than 2B parameters, which are both trained on a closed-source cross-embodiment datasets.
Despite their effectiveness, these approaches retain very large \glspl{vlm} with high-memory requirements and slow convergence. 
In contrast, \gls{flore} matches these methods with less than 1 billion parameters and significantly lower memory and fully open source training.

Efforts to reduce model size and finetuning overhead include using smaller \gls{vlm} backbones or alternative fusion strategies. 
TinyVLA attaches a compact diffusion head to a lightweight \gls{vlm} with a late-fusion approach and no pretraining \cite{wen2024tinyvla}. 
Fusion strategies range from early fusion, that merges raw \gls{vlm} outputs with the action generator at the input stage, to late fusion, where modalities are processed separately and combined only after independent streams \cite{wen2024tinyvla, liu2024rdt, hu2024videopredictionpolicygeneralist, li2024improving}. 
In contrast, intermediate fusion (our contribution) injects mid-level \gls{vlm} tokens into the Flow Transformer, allowing selective pruning of the \gls{vlm} while preserving semantic richness.
While DeerVLA \cite{yue2024deer} has explored early exit strategies for continuous action prediction \glspl{vla}, \gls{flore} significantly surpasses it across benchmarks by wide margins thanks to our novel flow-based fusion design.
Beyond model architecture, the choice of data sources and automated data collection are key to pretraining 
robust policies. 
Concurrent work has also explored intermediate-fusion \cite{bjorck2025gr00t} without any ablation and insights on the impact.
\gls{flore} advances efficient \glspl{vla} using intermediate-modality fusion and global-adaln conditioning, enabling faster training and inference while achieving SoTA performance on diverse benchmarks.

\section{Method }
\label{sec:method-overview}

We learn an efficient, generalist policy $\pi_\theta$ that generates actions conditioned on state $s_t$, text goal tokens $g_t$, and meta-embodiment information $e_i$. This task is challenging due to heterogeneity in action spaces (varying degrees-of-freedom, control modes), observation spaces (different sensors), and task specifications (diverse language goals). Given trajectories from multiple embodiments, we learn a unified policy that generalizes across robots, tasks, and observation formats by maximizing $\mathcal{L}_{\mathrm{IL}} = \mathbb{E}_{(s,a,g,e)\sim\mathcal{D}}[\log\pi_\theta(a\mid s,g,e)]$.

\subsection{Intermediate Modality Fusion Vision-Language-Action-Models}

On a high level, flow-based \gls{vla} consists of two main components: A \gls{vlm} pretrained on internet-scale vision and text data to encode state information and text and a flow prediction module to generate a sequence of actions conditioned on the current context from Gaussian noise. 

\begin{figure*}[t]
    \centering
    \includegraphics[width=1.0\textwidth]{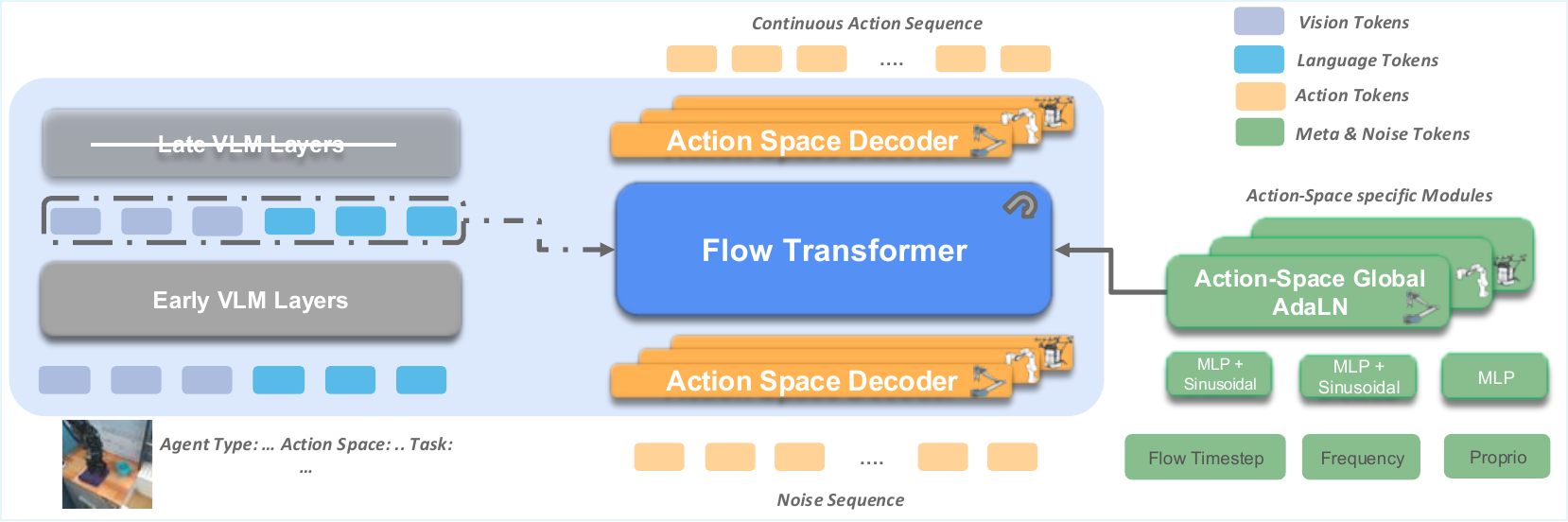}
    \caption{\textbf{\gls{flore} architecture.} A fine-tuned VLM processes multimodal inputs and integrates intermediate features into a Flow Transformer via cross-attention. The model predicts velocity fields using action-space Global AdaLN-Zero conditioning with embodiment and temporal metadata.
    }
    \label{fig:berg_arch}
\end{figure*}

We propose to fuse pretrained vision–language representations with Flow Transformer embeddings at an intermediate stage to balance semantic depth and computational efficiency. 
Prior work in \gls{llm} explainability shows that features from the penultimate quarter of transformer layers capture broad semantics, whereas final layers specialize in next-token prediction \cite{gao2024scaling}. 
Accordingly, this motivates our \gls{vla} fusion strategy to  extract hidden states from an intermediate layer of the \gls{vlm} backbone after jointly encoding visual (\gls{vit}) and textual tokens. 
These intermediate features preserve rich context without overspecialization for next-token prediction and reduced computational cost to enable more efficient \gls{vla} design.
We prune the \gls{vlm} based on its architecture: For \textbf{Encoder–Decoder VLMs} (Florence-2 \cite{xiao2024florence}), we remove the full decoder, keeping only the encoder \gls{llm} layers. This reduces the number of layers by $50\%$ while increasing its performance and efficiency.
For \textbf{Decoder-Only VLMs} (SmolFlow2-Video \cite{marafioti2025smolvlm}), we drop the final $30\%$ of transformer layers.
This targeted pruning cuts 20–35\% of parameters and reduces per-step latency. 
In \autoref{sec:evaluation_design_decisions}, we provide ablations to verify our intermediate fusion design. The full architecture is visualized in \autoref{fig:berg_arch}.

Next, we take the projected \gls{vlm} latent tokens and inject them into the Flow Transformer via cross‑attention. 
Specifically, we map the \gls{vlm} hidden states through a linear layer followed by RMSNorm \cite{zhang2019root} for increased stability. 
This design conditions each Flow layer on semantically rich \gls{vlm} features, preserving spatial and contextual structure and enabling faster policy convergence without the late fusion overhead.

\subsection{Cross-Action Space Flow Transformer}
\label{sec:cross-action-transformer}

\begin{wrapfigure}{r}{0.6\linewidth}
    \vspace{-\baselineskip}
    \centering
    \includegraphics[width=\linewidth]{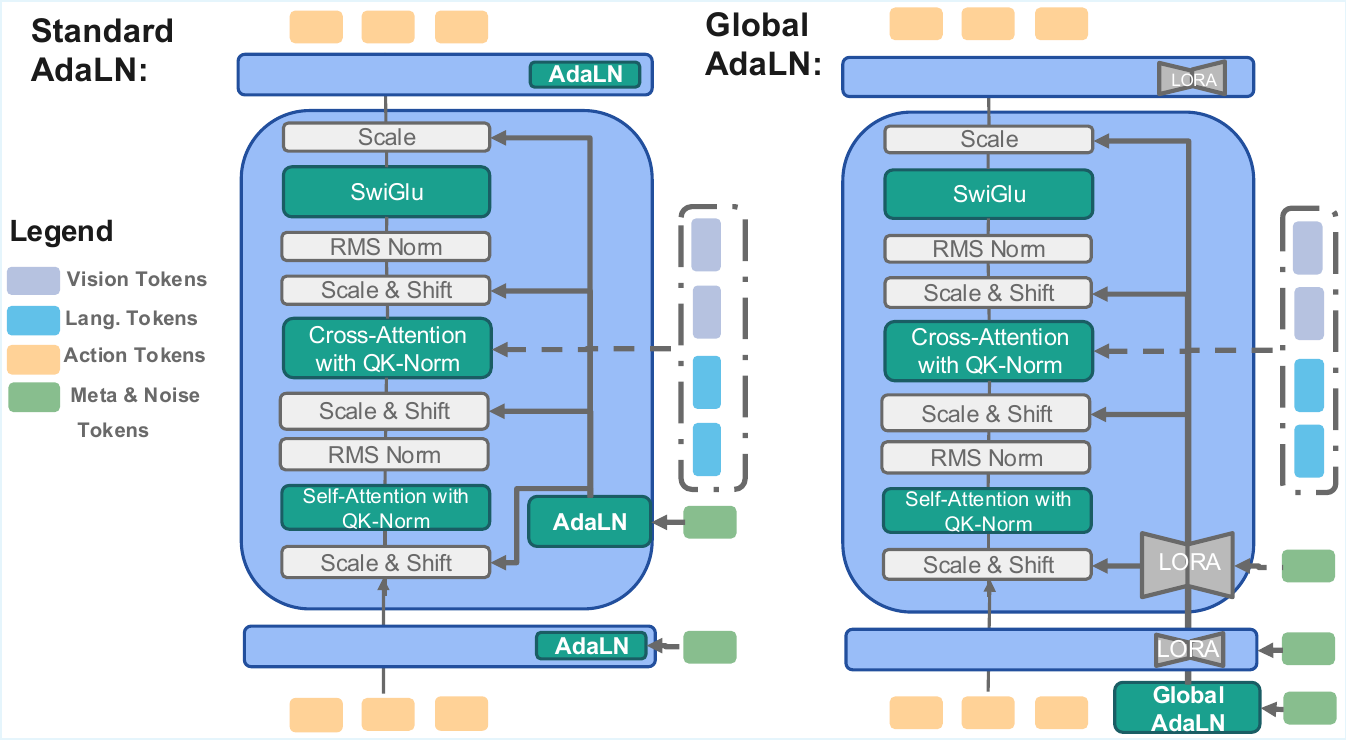}
    \caption{Comparison of standard DiT blocks and our proposed Global AdaLN with layer-specific Lora adapters.}
    \label{fig:adaln}
    \vspace{-\baselineskip}
\end{wrapfigure}
We design a novel, efficient Flow Transformer to handle heterogeneous action spaces efficiently.
In particular, we introduce \emph{Action-Space Global-AdaLN-Zero}, a unified normalization layer that conditions each transformer block on both temporal signals (e.g., flow time step) and per-action-type embeddings without incurring large parameter overheads, as illustrated in \autoref{fig:adaln}.

Standard AdaLN-Zero uses distinct scale-and-shift parameters per layer, adding up to $30\%$ extra parameters in diffusion transformers \cite{peebles2023scalable}.  
In contrast, \emph{Action-Space Global-AdaLN-Zero} shares a single set of modulation weights across all layers, while generating unique modulation signals for each action category (e.g., delta-EEF vs.\ joint angle), initialized with zeros for stable training.  
This reduces parameter count by over $20\%$ compared to naive AdaLN-Zero, yet preserves adaptation to action-space statistics.  
Given this lower per-layer expressiveness, we additionally inject lightweight LoRA adapters into each layer. This provides fine-grained, layer-specific modulation with only a few extra parameters per block.
Additionally, each action type uses a small encoder/decoder to map actions into/out of the transformer latent space, enabling consistent handling of differing action dimensions without sacrificing weight sharing.


\subsection{Rectified Flow for Action Generation}
\label{sec:rectified-flow}

We utilize Rectified Flow for generating actions. 
Flow models use straight-line velocity fields between noise and data distributions \cite{lipman2022flow, liu2022flow}. 
This approach reduces inference computation while maintaining expressiveness, which crucial for policies where latency matters.
For conditional action distribution $\p{\pi_\theta}{\act_{\itime,k} | \state_{\itime}, \goal, \emb}$, the trajectory interpolation follows:
\begin{equation}
    z_t = (1-t)\act_{\itime,k} + t z_1, \quad z_1 \sim \mathcal{N}(0, I),
\end{equation}
where $t \in [0, 1]$ is normalized flow time sampled from $t \sim \sigma(\mathcal{N}(0, I))$, $\act_{\itime,k} \in \mathbb{R}^{d_a}$ is the ground truth action sequence, and $z_1$ is standard Gaussian noise. The model optimizes:
\begin{equation}
    \mathcal{L}(\theta) = \mathbb{E}_{t, z_1} \left[ \| z_1 - \act_{\itime,k} - v_{\theta}(z_t, t, \state_{\itime}, \goal, \emb) \|^2 \right],
\end{equation}
where $v_\theta$ is the flow model conditioned on state $\state$, language goal $\goal$, and embodiment $\emb$. Inference requires only $N=4$ denoising steps for single-arm and $N=8$ for high-frequency dual-arm settings.



\subsection{FLOWER: Efficient Flow-based Vision-Language-Action Models}
\label{sec:flower}
Leveraging our contributions and insights in intermediate fusion and Global-AdaLN, we present \gls{flore}: a compact, efficient \gls{vla} policy with exceptional performance-to-parameter ratio. 
\gls{flore} uses half of the Florence-2-L \gls{vlm} as its primary backbone, which our ablations show provides optimal performance for robotic manipulation tasks. 
The model employs an $18$-layer Flow Transformer with $1024$ latent dimension.
Together with the action-specific modules, \gls{flore} has 947M parameters in total and only requires $1.85$ GB of VRAM.
This architecture achieves the computational efficiency required for real-time deployment

\textbf{Cost-Effective Pretraining Setup.}
To enable fast, cost-efficient pretraining, \gls{flore} uses a small, carefully chosen “OXE-soup” of eight public robotic datasets (approximately $250k$ trajectories total).  
We prioritize broad scene and embodiment diversity—including Franka Pandas and XARMs—and, inspired by \citet{lin2024data}. $75\%$ of our data samples from Droid \cite{khazatsky2024droid}, Google Robot \cite{brohan2023rt}, and BridgeV2 \cite{walke2023bridgedata}. Unlike most OXE datasets, these data are gathered in varied environments with rich distractions, backgrounds, and objects. We use an action chunk length of $20$ and a single static image input \cite{gao2024scaling,liu2024ok}.
By training on a compact mixture of $74\%$ delta-EEF and $26\%$ single-arm joint-state data, we complete $360,000$ steps in 48 h ($\approx 200$ GPU-hours).
Extending the run yields no further gains (see App. \ref{sec:pretraining-details}).
To identify this optimum, we track zero-shot success on the Bridge SIMPLER benchmark \cite{li24simpler}, observing that \gls{flore} Bridge performance stagnates thereafter.

\label{sec:sim-experiments}
\begin{figure*}[t]
    \centering
    \includegraphics[width=\linewidth]{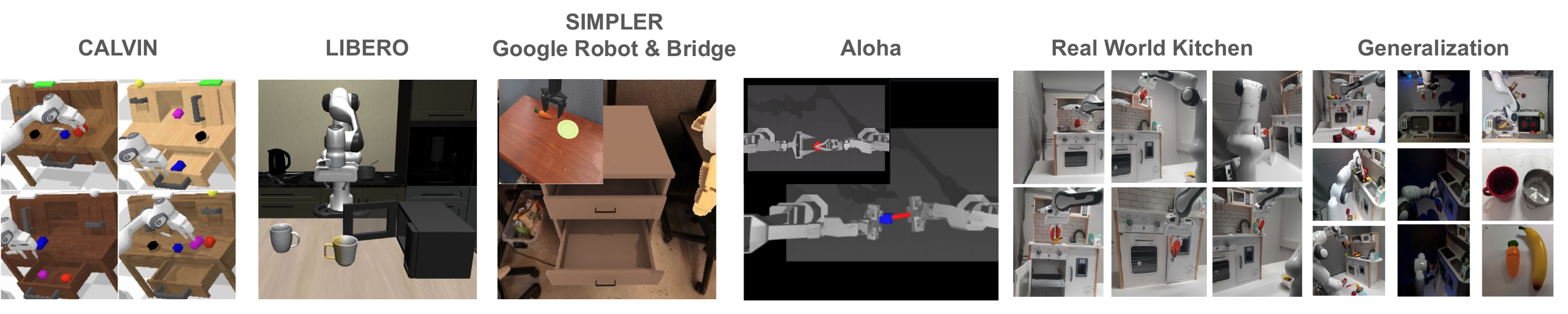}
    \caption{\textbf{Simulation Environments used to test \gls{flore}.} From left to right: \textit{CALVIN}~\cite{mees2022calvin}, \textit{LIBERO}~\cite{liu2024libero}, \textit{SIMPLER}~\cite{li24simpler} with the \textit{Bridge} and \textit{Google Robot} variants and \textit{Aloha Simulation Benchmark}~\cite{zhao2023learning}. Real world multi-task kitchen setup and generalization experiments with cluttered scenes, different lightning and novel objects.}
    \label{fig:environments}
\end{figure*}

\section{Evaluation}
\label{sec:evaluation}

We aim to answer the following research questions in our experiments:
\textbf{(RQ I)} How do the proposed design elements of our novel \gls{vla} architecture impact performance?
\textbf{(RQ II)} Does our \gls{vla}-design deliver strong performance while significantly reducing computational demands compared to sota \gls{vla} policies across diverse settings?
\textbf{(RQ III)} Can we train a \gls{vla} policy with less than 1B parameters that handles robot embodiments (e.\,g., single-arm vs.\ dual-arm, delta-EEF vs.\ joint angles) and robustly transfers to unseen environments, novel objects, and varying conditions?
We evaluate our method across more than 190 tasks in 10 different benchmarks to answer these questions.

\subsection{Evaluation of Critical Design Decisions for Efficient Flow VLAs}
\label{sec:evaluation_design_decisions}


\begin{table}
\setlength{\tabcolsep}{3.5pt} 
\begin{center}
\parbox[t]{0.65\textwidth}{%
    \footnotesize
    \begin{tabular}{|l|cc|cc|}
    \hline
    \multirow{3}{*}{\makecell[l]{Fusion\\Strategy}} & \multicolumn{4}{c|}{Success Rate (\%)} \\
    \cline{2-5}
     & \multicolumn{2}{c|}{Florence-VLM} & \multicolumn{2}{c|}{SMol-VLM} \\
    \cline{2-5}
     & C-ABC & L-Long & C-ABC & L-Long \\
    \hline
    Early    & $57.1 \pm 5.3$ & $33.4 \pm 6.0$ & $25.8 \pm 3.9$ & $44.5 \pm 2.7$ \\
    Inter.  & $\mathbf{89.5 \pm 1.0}$ & $\mathbf{93.4 \pm 2.0}$ & $\mathbf{72.1 \pm 5.0}$ & $\mathbf{70.7 \pm 2.3}$ \\
    Late     & $71.2 \pm 2.2$ & $61.8 \pm 2.5$ & $66.3 \pm 2.0$ & $69.2 \pm 1.9$ \\
    \hline
    \end{tabular}
    \captionsetup{width=0.65\textwidth}
\caption{\textbf{Evaluation of different \gls{vla} fusion strategies.}\newline 
Intermediate fusion yields the best performance across both\newline \gls{vlm} types.}

    \label{tab:fusion_ablation}
}%
\hfill
\parbox[t]{0.33\textwidth}{%
    \footnotesize
    \setlength{\tabcolsep}{3.5pt} 
    \begin{tabular}{|l|cc|}
    \hline
    \multirow{2}{*}{\makecell[l]{Layers}} & \multicolumn{2}{c|}{Model} \\
    \cline{2-3}
     & C-ABC & L-Long \\
    \hline
    Full & $66.3 \pm 2.0$ & $69.2 \pm 1.9$ \\
    \hline
    0.2 & $68.6 \pm 3.2$ & \textbf{$71.8 \pm 3.7$} \\
    \hline
    0.3 & \textbf{$72.1 \pm 5.0$} & $70.7 \pm 2.3$ \\
    \hline
    0.5 & $66.4 \pm 6.4$ & $62.5 \pm 3.5$ \\
    \hline
    \end{tabular}
    \caption{\textbf{Comparing Layer Pruning} for CALVIN ABC and LIBERO Long.}
    \label{tab:pruning_comparison}
}
\end{center}
    \vspace{-2.5\baselineskip}
\end{table}

In this section, we evaluate the key design decisions that impact the efficiency and performance of small Flow-based \glspl{vla} across two benchmarks: CALVIN ABC \cite{mees2022calvin} and LIBERO-Long \cite{liu2024libero}. We train each policy variant with $3$ seeds for up to $100k$ steps and report the average performance.
While CALVIN-ABC focuses on generalization for free form instruction following to solve $34$ different task, LIBERO-Long emphasizes long-horizon task completion. 
Both benchmarks are established for testing \gls{vla}s and provide useful insights into optimal design. 

\textbf{Does intermediate fusion provide strong performance with higher efficiency?} The fusion strategy determines how the \gls{vlm}'s vision-language features integrate with the action prediction model. We compare three strategies: early fusion (combining noise token with \gls{vlm} tokens at \gls{llm} level), our proposed intermediate fusion, and late fusion (using final \gls{vlm} outputs to condition the flow prediction module). 
As shown in Table~\ref{tab:fusion_ablation} and Table~\ref{tab:pruning_comparison}, intermediate fusion outperforms alternatives with both model variants and on both benchmarks. 
With Florence-VLM, it achieves $93.4\%$ success on LIBERO-Long, a $61$ percentage point improvement over early fusion ($33.4\%$) and $21$ points over late fusion ($73\%$). 
This confirms our hypothesis that intermediate-layer features provide an optimal balance of semantic richness and computational efficiency, addressing \textbf{(RQ I)}.

\begin{wraptable}{r}{0.3\textwidth}
\vspace{-\baselineskip}
\centering
\resizebox{0.3\textwidth}{!}{%
\centering
\begin{tabular}{l|c}
\toprule
\textbf{Variant} & \textbf{Avg. Len.} \\
\midrule
FLOWER & \textbf{4.44}$\pm$\textbf{0.04} \\
+ standard AdaLN & 4.43$\pm$0.03 \\
- Flow Head & 3.33$\pm$0.04 \\
- Custom LR & 4.40$\pm$0.05 \\
- No VLM train & 2.65$\pm$0.36 \\
+ small Florence & 4.26$\pm$0.04 \\
- VLM & 3.42$\pm$0.07 \\
+ Discrete Token & 1.12$\pm$0.12 \\
+ Smaller Head & 2.60 $\pm$0.09 \\
\bottomrule
\end{tabular}
}
\caption{Average Sequence Lengths for FLOWER Ablations on CALVIN ABC. }
\label{tab:FLOWER-Ablations}
\vspace{-\baselineskip}
\end{wraptable}

\textbf{What type of VLM backbone is best for efficient VLAs?} We explore different \gls{vlm} backbones, comparing Florence-2-L \cite{xiao2024florence} and SmolFlow-500M \cite{marafioti2025smolvlm}. 
Beyond their architectural differences (Florence-2 being encoder-decoder and SmolFlow decoder-only), their pretraining objectives significantly differ. Florence-2 was trained on FLD-5B (5.4 billion annotations across 126 million images) with emphasis on object detection, segmentation and visual grounding, while SmolFlow's pretraining  prioritizes general reasoning and language understanding capabilities across text, videos and images. 
This is representative for a standard \gls{vlm} pretraining. 
Our experiments confirm that Florence-2's pretraining translates more effectively to robotic manipulation tasks, further addressing \textbf{(RQ I)}. 
We therefore use the Florence-2 variant of \gls{flore} for all subsequent experiments. 

\textbf{Does Global-AdaLN enable more efficient Flow Transformers?}
Next, we compare our proposed Global-AdaLN against default AdaLN used in prior work \cite{reuss2024multimodal, ke20243d} for our Florence-based \gls{flore} model on CALVIN ABC. 
The results in \autoref{tab:FLOWER-Ablations} demonstrate that our proposed Global AdaLN enables relevant parameter reduction of $20\%$ without reducing the performance. 

\textbf{Do we need a large-capacity Diffusion Transformer?}
Next, we test \gls{flore} with a small action head using $384$ latent dimensions and 6 layers. The final model only achieves significantly lower average performance of $2.60$ compared to our design. 
This confirms that having a high capacity Diffusion Transformer is crucial for performance.

\textbf{What other design choices matter?}
Finally, we test our \gls{vla} with several ablations variants using the Florence2 version on CALVIN ABC (\autoref{tab:FLOWER-Ablations}). 
The smaller Florence reduces performance notably, while the frozen \gls{vlm} has an even bigger negative impact on performance. 
In addition, our custom Learning Rate scheduler increases performance.
Finally, we compare \gls{flore} with an L1-prediction head like ACT \cite{zhao2023learning} and a discrete token variant to demonstrate the importance of using flow prediction. These findings complete our analysis for \textbf{(RQ I)}.

\subsection{Simulation Experiments}

\begin{figure*}[t]
    \centering
    \includegraphics[width=1\linewidth]{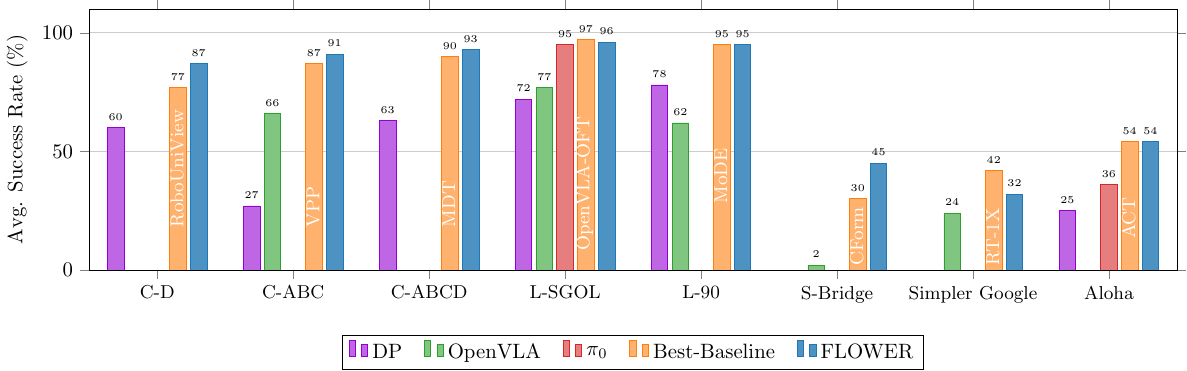}
    \caption{\textbf{Simulation Results for FLOWER} We report average results for various benchmarks against relevant baselines. For brevity we reduce the shown baselines to most relevant ones but provide detailed results for each benchmark (see \autoref{sec:app-benchmarks}. C refers to CALVIN and L refers to LIBERO. SGOL refers to average results for LIBERO Object, Goal, Spatial and Long.}
    \label{fig:sim-all}
    \vspace{-1.5\baselineskip}
\end{figure*}

Next, we take the best performing \gls{vla} variant, \gls{flore}, with the Florence-2-L backbone and pretrain it on our pretraining mix. After pretraining, we evaluate \gls{flore} across multiple simulation benchmarks to assess its performance, generalization capabilities, and adaptation to different robotic environments.

\textbf{Benchmark Evaluation.}
We evaluate FLOWER on four established benchmark suites representing diverse robotic manipulation challenges:
\textbf{CALVIN} \cite{mees2022calvin} features 34 tabletop manipulation tasks with a Franka Panda robot using delta end-effector control across four scene configurations (splits A-D). The dataset contains 24,000 language-annotated demonstrations. 
We evaluate three settings: CALVIN D, CALVIN ABC (zero-shot generalization), and CALVIN ABCD (scaling with more data). 
Performance is measured by success rates on sequential tasks and mean sequence length completion. All evaluations require policies to follow free-form language instructions and complete 5 tasks in sequence across 1,000 different instruction chains.
\textbf{LIBERO} \cite{liu2024libero} tests a delta-EEF controlled Panda Robot across various scenes and challenges. 
We report results on four specialized variants with 10 tasks each (Long, Spatial, Object, and Goal) and separately on LIBERO-90, which requires policies to solve 90 different tasks in diverse scenes. Success is measured as the percentage of successful task completions across 50 trials per task (20 for LIBERO-90).
\textbf{Aloha} \cite{zhao2023learning} simulation tasks evaluate bi-manual joint state manipulation, requiring high-frequency control (50Hz), where tasks include cube transfer and peg insertion.
\textbf{SIMPLER} \cite{li24simpler} provides  evaluation after pretraining on real2sim environments with Bridge and Google Robot variants. 

\textbf{Baselines.}
We compare our \gls{vla} against sota \gls{vla} policies and specialized approaches, using results reported in prior publications for fair comparison. Our primary comparisons are with OpenVLA \cite{kim2024openvla} (7.7B parameters), $\pi_0$ \cite{black2024pi_0} (3.3B parameters), and a standard Diffusion Policy using a CNN \cite{chi2023diffusionpolicy}. 
We provide detailed comparisons against additional relevant baselines for each benchmark in Appendix \ref{sec:app-benchmarks}, including video-based \glspl{vla} like VPP \cite{hu2024videopredictionpolicygeneralist}, and specialized policies like Baku \cite{haldar2024baku}.

\textbf{Results.}
The results for these experiments are summarized in \autoref{fig:sim-all}.
\gls{flore} consistently matches or surpasses all current SoTA approaches across all reported benchmarks. 
Notably, it surpasses OpenVLA on CALVIN and LIBERO by wide margins despite its smaller size and faster training. It also outperforms $\pi_0$ on the LIBERO and ALOHA benchmarks. 
Overall, these results demonstrate the versatility of  \gls{flore} to adapt do diverse embodiments and task settings.
These results confirm that \gls{flore} provides strong performance at low computational cost \textbf{(RQ II)} and demonstrates generalization to unseen environment settings \textbf{(RQ III)}. 

\begin{figure*}[t]
    \begin{subfigure}{0.48\linewidth}
        \centering
        \includegraphics[width=\linewidth]{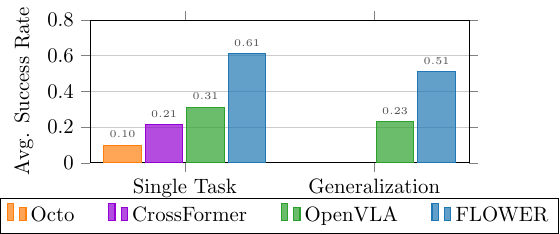}
        \caption{\textbf{Real World Results} for different Generalist Policies finetuned on a Franka Panda Kitchen Setup.}
        \label{fig:real-robot-results-summary}
    \end{subfigure}
    \hfill
    \begin{subfigure}{0.48\linewidth}
        \centering
        \scalebox{0.85}{
        \begin{tabular}{lcc}
        \toprule
        \textbf{Generalization Scenario} & \textbf{FLOWER} & \textbf{OpenVLA} \\
        \midrule
        Novel Object       & \textbf{33.3\%}   & 10.0\% \\
        Flashlight         & \textbf{50.0\%}   & 25.0\% \\
        BG Distractors     & \textbf{69.5\%}   & 41.7\% \\
        New Tasks Composition & \textbf{51.1\%}   & 16.7\% \\
        \midrule
        Average            & \textbf{51.0\%}   & 23.4\% \\
        \bottomrule
        \end{tabular}
        }
        \caption{\textbf{Generalization Results} comparing FLOWER and OpenVLA across different scenarios.}
        \label{tab:generalization_summary}
    \end{subfigure}
    \caption{\textbf{Real World Results for Multi-Task and Generalization Experiments}}
    \label{fig:combined-results}
\end{figure*}

\subsection{Real-World Evaluation and Generalization}

Next, we want to verify our results from simulation in a real world setup. 
Therefore, we evaluate \gls{flore} in a real-world kitchen setting with a Franka Panda Robot across $20$ distinct tasks involving various objects. 
The training data consists of $417$ language-annotated trajectories from $45$ minutes of human demonstrations collected via kinesthetic teaching. 
All Policies are finetuned in joint state space at $6$ Hz.
An overview of various tasks is shown in \autoref{fig:environments} and we additionally provide rollouts in our project page.
We compare \gls{flore} against several generalist VLA policies: Octo \cite{octo_2023}, OpenVLA \cite{kim2024openvla}, and CrossFormer \cite{doshiscaling}, all finetuned on our dataset using recommended hyperparameters (see \autoref{subsec:finetuning-baselines-real-world} for details).

\textbf{Results.} First, we test all policies on instruction following for all 20 different tasks. 
Each task is evaluated five times from randomized starting positions. 
As shown in \autoref{fig:real-robot-results-summary}, \gls{flore} achieves the highest average success rate ($61\%$), doubling the performance of the second-best baseline, OpenVLA ($31\%$). 
These results address \textbf{(RQ II)} by demonstrating \gls{flore}'s strong performance compared to state-of-the-art baselines in real-world settings.

\textbf{Generalization Analysis.} We further test \gls{flore}'s generalization capabilities against OpenVLA under challenging conditions: novel objects, flashlight-only lighting, background distractors, and new task compositions. 
As summarized in \autoref{tab:generalization_summary}, \gls{flore} consistently outperforms OpenVLA across all scenarios, averaging $51.0\%$ success compared to OpenVLA's $23.4\%$. 
These experiments verify that \gls{flore} can effectively adapt to unstructured, real-world variations \textbf{(RQ III)}, though it still faces challenges with fine manipulation in highly cluttered environments.

\begin{wraptable}{r}{0.5\textwidth}
\vspace{-\baselineskip}
\centering
\resizebox{0.5\textwidth}{!}{%
\centering
\begin{tabular}{lccc}
\hline
Method & Throughput (Hz)$\uparrow$ & Latency (s)$\downarrow$ & VRAM (MB)$\downarrow$ \\
\hline
DP (0.26B) & 130.67 & 0.341 & \textbf{517} \\
OpenVLA (7B) & 6.09 & 0.164 & 14574 \\
$\pi_0$ (3.3B) & \textit{288.11} & \textit{0.104} & 6692 \\
\textbf{FLOWER+LF(1.15B)} & 287.36 & 0.055 & 2235 \\
\textbf{FLOWER (0.95B)} & \textbf{311.04} & \textbf{0.052} & \textit{1848} \\
\hline
\end{tabular}
}
\vspace{-0.25em}
\caption{\textbf{Mean inference efficiency} (1000 steps in Bf16). 
All policies except OpenVLA use chunk length $50$. \textit{LF} refers to Late Fusion Ablation of FLOWER with the complete VLM.}
\label{tab:inference_efficiency}
\vspace{-\baselineskip}
\end{wraptable}
\textbf{Inference Efficiency.} 
We evaluated \gls{flore} against baselines on an RTX 4090 GPU (Table~\ref{tab:inference_efficiency}). 
\gls{flore} achieves a throughput of $311$Hz, making it 8\% faster than $\pi_0$ ($288$Hz), and $5007$\% faster than OpenVLA ($6.1$Hz). 
\gls{flore} has \textbf{the lowest memory footprint} among VLAs, using $27.6$\% of $\pi_0$'s memory and $12.7$\% of OpenVLA's, making it ideal for commodity hardware. These results strongly support \textbf{(RQ II)}.


\section{Conclusion}
\label{sec:conclusion}
We introduced several contributions for efficient \glspl{vla}: intermediate-level fusion that preserves semantic understanding while pruning 30-50\% of VLM layers, and global action-space AdaLN that reduces transformer head parameters by 20\% without compromising expressivity. These techniques enable compact yet powerful flow-based \glspl{vla}.
This yields \gls{flore}, an efficient \gls{vla} that matches current state-of-the-art VLAs across 190 tasks in 10 benchmarks, despite requiring only 950M parameters and 200 GPU hours for pretraining. 

\section{Limitations}
\label{sec:limitations}
Despite its advantages, \gls{flore} still has several limitations.
First, it relies on an iterative sampling procedure, which is inherently slower than a single forward pass from deterministic policies. 
Second, we have validated \gls{flore} primarily on three manipulation action spaces; its ability to generalize to other embodiments, such as mobile navigation or humanoid locomotion, remains unexplored and is an important direction for future work. 
Third, pretraining performance for zero-shot deployment on the SIMPLER Google Robot benchmark indicates that further improvements are needed. 
We hypothesize that the generalization tested in SIMPLER benefits from larger models. 
Fourth, although \gls{flore} is considerably smaller than most state-of-the-art \gls{vla} models, its $\approx1$ B-parameter size may still present deployment challenges in low-resource or real-time settings.
Fifth, eight out of our ten used benchmarks are conducted in simulation, limiting the extent to which our results can be taken as evidence of real-world generalization.

\section{Acknowledgments}

The work was funded by the German Research Foundation (DFG) – 448648559. 
The authors also acknowledge support by the state of Baden-Württemberg through HoreKa supercomputer funded by the Ministry of Science, Research and the
Arts Baden-Württemberg and by the German Federal Ministry of Education and Research.

\bibliography{references}  
\newpage

\appendix

\begin{table*}[]
\scalebox{0.55}{
    \centering
    \begin{tabular}{c|ccccc}
         & SIMPLER & CALVIN & LIBERO & Aloha & Real World Kitchen  \\
         \midrule
    Action Space Encoders & \multicolumn{5}{c}{2-layered MLP} \\ 
    Action Space Decoders & \multicolumn{5}{c}{Linear Attention} \\ 
    Number of Flow-T Layers & \multicolumn{5}{c}{18} \\
    Latent Dimension & \multicolumn{5}{c}{1024} \\ 
    Number of Heads & \multicolumn{5}{c}{16} \\ 
    Position Embedding &  \multicolumn{5}{c}{1D Rope} \\ 
    Sampling Distribution & \multicolumn{5}{c}{Uniform} \\
    Attention Dropout & \multicolumn{5}{c}{0.1} \\ 
    MLP Dropout & \multicolumn{5}{c}{0.1} \\ 
    Residual Dropout & \multicolumn{5}{c}{0.1} \\ 
    \midrule
     Act Seq Length    & 10 & 10  & 10 & 100 & 20 \\
     Denoising Steps & 4 & 4 & 4 & 8 & 5 \\
     Multistep    & 5 & 10 & 10 & 2: Insert 100:Transfer & 15 \\
     Camera Views  & [Primary Static] & [Primary Static, Wrist] & [Primary Static, Wrist] & [Primary Static] & [Primary Static, Secondary Static]\\
     Use Proprio  & False & False & False & True & False \\
     Action Space & Delta EEF & Delta EEF & Delta EEF & Bi-Joint & Joint \\
     Frequency & 3/5 & 10 & 10 & 50 & 6 \\
     \bottomrule
    \end{tabular}
    }
    \caption{\textbf{Overview of Hyperparameters used for FLOWER across different Benchmarks}}
    \label{tab:my_label}
\end{table*}

\begin{table}[h]
    \centering
    \begin{tabular}{l|c}
         \textbf{Name} & \textbf{Number of Parameters} \\
        \hline
        ViT  & 360M  \\
        VLM  & 205M  \\
        Action Encoders  & 3.2M  \\
        Action Heads   & 31.8K \\
        Global-AdaLN   & 28.3M \\
        Cond Linear Proj. & 1.0M  \\
        Timestep Embedder  & 1.3M  \\
        Cond Norm   & 1.0K  \\
        FreqEmbedder  & 1.3M  \\
        Flow Transformer  & 339M  \\
        \midrule 
        Total Parameters FLOWER & 947M \\
        \bottomrule
    \end{tabular}
    \caption{\textbf{Overview of Parameter Distribution across all Model Components of FLOWER.}}
    \label{tab:model_components-parameteres}
\end{table}

\section{Pretraining Details}
\label{sec:pretraining-details}

\begin{table}[htbp]
\centering
\caption{Dataset Distribution by Percentage}
\begin{tabular}{lc}
\hline
\textbf{Dataset} & \textbf{Percentage (\%)} \\
\hline
bridge\_dataset & 28.62 \\
fractal\_data & 24.68 \\
droid & 23.50 \\
cmu\_play\_fusion & 6.15 \\
dobbe & 5.94 \\
libero\_10\_no\_noops & 4.41 \\
libero\_goal\_no\_noops & 4.07 \\
real\_kitchen\_lang & 2.64 \\
\hline
\end{tabular}
\label{tab:dataset_percentages}
\end{table}

\begin{table}[]
    \centering
    \begin{tabular}{c|c}
    \textbf{Hyperparameter} & \textbf{Value} \\
    \midrule 
    GPU Type & H100 \\
    N GPUS & 4 \\
    Batch Size & 256 \\
    Grad accumulate Steps &  4 \\
      Optimizer FlowT   &  AdamW \\
      LR Max FlowT & 1e-4 \\
    LR Scheduler  & warm-up with constant + cosine decay \\
    Min LR FlowT  & 1e-5\\
    Final LR FlowT  & 1e-5 \\
    FlowT LR scheduler phases & [0.01, 0.39, 0.6] \\
    Max  Desired Train Steps & 600,000 \\
     Optimizer VLM   &  AdamW \\
      Lr Max VLM  & 1e-5 \\
    Lr Scheduler  & warm-up with constant + cosine decay \\
    Min LR VLM   &  1e-7 \\
    Final LR VLM   & 1e-6 \\
    VLM  LR scheduler phases & [0,1, 0.3, 0.6] \\
    EMA & False \\
    Weight Decay FlowT & 0.1 \\
    Weight Decay VLM & 0.001 \\
    Total Training Time & 48 hours \\
    Training Steps reached & 350,000 \\
    \bottomrule
    \end{tabular}
    \caption{\textbf{Hyperparameteres for Pretraining FLOWER on all Pretraining Data Mixtures}}
    \label{tab:pretraining-hyperparameters}
\end{table}

We pretrain \gls{flore} on different datasets mixes that are described in \autoref{tab:dataset_percentages} using one cluster node with 4 H100 GPUs for 48 hours.
We pretrain all variants using a single static image and only use proprioception signal for the bimanual settings. 
We set the action chunk length to 20 across all settings and condition the model on single image from the current state only for maximum efficiency. 
Training is conducted using HuggingFace Accelerate for optimized Multi-GPU Training. 
We created custom PyTorch wrapper for the OXE Torch Dataloaders \cite{open_x_embodiment_rt_x_2023} inspired by efforts from OpenVLA \cite{kim2024openvla}.
We train \gls{flore} using BF-16 accuracy for optimized memory performance. 
For the Cross-Action-and Delta-EEF Mix we set the batch size to 256 and use 4 gradient accumulate steps to achieve an batch size of 1024. 
In total we trained for 300k-400k training steps depending on the dataset composition. 
We did not notice major training instabilities. 

\subsection{Pretraining Ablation Experiences}

We tested several ideas, that did not work well in our maximum efficiency pretraining settings:
\begin{itemize}
    \item \textbf{Using variable Length Action Chunks:} We experimented with flexible action chunks during training as done in CrossFormer \cite{doshiscaling}. However, we noticed slow convergence and lower performance for our SIMPLER experiments. 
    Thus, we decided to use a constant chunk length for all datasets of 20. While many delta EEF datasets like Fractal operate on lowe frequency and Aloha Setups like Biplay \cite{dasari2024ingredients} require 50 Hz, we find 20 to be a good trade-off that enables easy finetuning to different lengths, thanks to our 1D Rope Position Embeddings.

    \item \textbf{Using Multiple Images with Custom Masking:} We also tested variations in pretraining that involve flexible image padding up to 3 different views depending on the dataset. However, this reduces the overall training speed given the higher number of image tokens and required GPU memory. Also given the more complex setting, we found that the training time of 2 days not enough to achieve convergence. Thus, we limited pretraining to single image. However, our finetuning experiments show the flexbility of \gls{flore} to adapt to more image views without issues.

    \item \textbf{Mixture-of-Experts Approaches for the Flow Transformer}: We conducted architecture ablation experiments with more action-specialist components. \gls{flore} ablations used action-type specific MLPs with action-type specific LayerNorms. We also experimented with a shared additional MLP, that all action types use while combining the output with action-specific specialist MLP. However, the training was more memory intensive and had issues with NaN losses. It also showed slower convergence. We found the action-specific Global-AdaLN with all other parameters shared inside the Flow Transformer to work best. However, we believe that future research can address this issue to develop even more efficient architectures for cross-embodiment learning.

\end{itemize}

\subsection{Language Prompt for the VLA}

We encode the meta-information as part of our prompt for the VLM: 
\textit{``Agent Type: [robot\_type], Action Space: [action\_space], Task: [task\_description]''.}

\subsection{Custom Learning Rate Scheduler}

In early experiments with an older version of our model, we observed a substantial drop in performance when training without our custom dual–optimizer learning rate scheduler (4.44 vs 0.8).
Upon further investigation after finalizing the model, we found that this was largely due to instabilities in an earlier model variant rather than the absence of the scheduler itself. 
With our final architecture, replacing the custom scheduler with a standard constant learning rate of $2 \times 10^{-5}$ leads to only a slight reduction in performance, confirming that the scheduler provides only minor benefits rather than being a critical requirement for the performance.

\subsection{Details for Cross-Action Space Flow Transformer}

Large transformer models often face stability challenges when simultaneously dealing with different data frequencies and distributions. 
We address this by allocating individual dual-RMSNorm parameters for each action type, capturing action-specific activation statistics more effectively than a single normalization. 

Additionally, we replace standard feed-forward layers with SwiGlu blocks~\cite{shazeer2020glu}, a sandwich-style MLP that we express as follows. 
For an input vector $\mathbf{h}$,
\begin{equation}
\label{eq:swiglue}
\mathbf{y} \;=\; \Bigl(\mathrm{Norm}(\mathbf{W}_1\,\mathbf{h})\Bigr) \;\odot\; \mathrm{SiLU}\!\Bigl(\mathrm{Norm}(\mathbf{V}_1\,\mathbf{h})\Bigr),
\end{equation}
where $\mathrm{SiLU}(x) = x \,\sigma(x)$, $\mathbf{V}$ and $\mathbf{W}$ are to linear matrices and $\mathrm{Norm}(\cdot)$ indicates RMSNorm in our implementation.
Unlike LayerNorm, which subtracts the mean, RMSNorm normalizes each sample by its root-mean-square:
\begin{equation}
\mathrm{RMSNorm}(\mathbf{x}) \;=\; \frac{\mathbf{x}}{\sqrt{\tfrac{1}{d}\sum_{j=1}^d x_j^2 \;+\;\epsilon}},
\end{equation}
yielding smoother gradients and reduced training instability~\cite{zhang2019root}. Furthermore, we apply QK-value normalization~\cite{henry2020query} in both self- and cross-attention modules to mitigate large softmax outputs.
This additional normalization has been used in many large-scale diffusion and flow transformer architectures in image generation \cite{esser2024scaling} or Diffusion Policies \cite{liu2024rdt}.

Together, (i) extended RoPE embeddings, (ii) action-specific encoders/decoders, (iii) a shared AdaLN controller yielding action-specific normalization signals, and (iv) RMSNorm with SwiGLU MLPs enable our Flow Transformer to train stably across heterogeneous data and multiple action spaces. Its modular architecture requires minimal changes when adding new action types, retaining efficient scalability for a wide range of robotic tasks.

\section{Detailed Experiments}
\label{sec:app-benchmarks}

In this section we provide detailed results and additional comparisons for all simulation environments and real robot experiments. 

\begin{table}[ht]
\centering
\scalebox{1.0}{
\begin{tabular}{lcccc}
\toprule
\textbf{Benchmark} & \textbf{FLOWER} & \textbf{2nd Best} & \textbf{Abs. Imp.} & \textbf{Rel. Imp. (\%)} \\
\midrule
CALVIN D              & 87.0\% & 77.0\% & 10.0\%  & 13.0\%   \\
CALVIN ABC            & 90.6\% & 85.8\% & 4.8\%   & 5.6\%    \\
CALVIN ABCD           & 93.4\% & 90.4\% & 3.0\%   & 3.3\%    \\
SIMPLER Bridge        & 40.0\% & 30.0\% & 10.0\%  & 33.3\%   \\
SIMPLER Google        & 32.2\% & 42.4\% & -10.2\% & -24.1\%  \\
LIBERO OLGS           & 96.9\% & 97.1\% & -0.2\%  & -0.2\%   \\
LIBERO 90             & 94.7\% & 96.0\% & -1.3\%  & -1.4\%   \\
Real-World            & 61.0\% & 30.0\% & 31.0\%  & 103.3\%  \\
Aloha                 & 54.0\% & 54.0\% & 0.0\%   & 0.0\%    \\
Real-Generalization   & 51.0\% & 23.4\% & 27.6\%  & 118.0\%  \\
\midrule
\textbf{Average}      & --     & --     & 7.5\%   & 25.1\%   \\
\bottomrule
\end{tabular}}
\caption{Normalized performance improvement of FLOWER compared to its second-best baseline for each benchmark. CALVIN metrics are normalized by dividing the average sequence lengths by 5. Real-Generalization values are computed as the average across Novel Object, Flashlight, BG Distractors, and New Tasks Composition tests. The overall average improvement is computed over all benchmarks.}
\label{tab:flower_improvements}
\end{table}

\begin{table*}
\centering
\scalebox{0.75}{
\begin{tabular}{lll|l|l|cccccc}
\toprule
Train$\rightarrow$Test & Method & PrT & Action Type & VLM & \multicolumn{5}{c|}{No. Instructions in a Row (1000 chains)} & Avg. Len. \\  
\cmidrule(lr){5-9}
&            &     &            &     & 1    & 2    & 3    & 4    & 5    &   \\  
\midrule
\multirow{8}{*}{ABC$\rightarrow$D} &Diff-P-CNN  \cite{chi2023diffusionpolicy}   & $\times$      & Diffusion  & $\times$ & 63.5\% & 35.3\% & 19.4\% & 10.7\% & 6.4\%  & 1.35$\pm$0.05 \\  
 & MDT   \cite{reuss2024multimodal}   & $\times$      & Diffusion  & $\times$ & 63.1\% & 42.9\% & 24.7\% & 15.1\% & 9.1\%  & 1.55 \\  
&RoboFlamingo \cite{li2023vision}  & $\times$  & Cont.  & $\checkmark$ & 82.4\% & 61.9\% & 46.6\% & 33.1\% & 23.5\% & 2.47 \\ 
& SuSIE   \cite{black2023zero}      & $\checkmark$  & Diffusion & $\times$ & 87.0\% & 69.0\% & 49.0\% & 38.0\% & 26.0\% & 2.69 \\
&DeerVLA \cite{yue2024deer}  & $\times$  & Cont.  & $\checkmark$ & 84.8\% & 72.3\% & 54.9\% & 44.6\% & 33.5\% & 2.90 \\ 
&GR-1 \cite{wu2023unleashing}        & $\checkmark$  & Cont. & $\times$ & 85.4\% & 71.2\% & 59.6\% & 49.7\% & 40.1\% & 3.06 \\
&OpenVLA \cite{kim2024openvla}          & $\checkmark$      & Discrete & $\checkmark$ & 91.3\% & 77.8\% & 62.0\% & 52.1\% & 43.5\% & 3.27 \\ 
&3DDA \cite{ke2024d}          & $\times$      & Diffusion & $\times$ & 93.8\% & 80.3\% & 66.2\% & 53.3\% & 41.2\% & 3.35 \\ 
& MoDE \cite{reuss2024efficient}           & $\checkmark$  & Diffusion & $\times$ & 96.2\% & 88.9\% & 81.1\% & 71.8\% & 63.5\% & 4.01$\pm$0.04 \\  
& RoboDual \cite{bu2024towards}          & $\checkmark$      & Diffusion & $\checkmark$ & 94.4\% & 82.7\% & 72.1\% & 62.4\% & 54.4\% & 3.66 \\ 
& VPP  \cite{hu2024videopredictionpolicygeneralist}         & $\checkmark$  & Diffusion & $\times$ & 95.7\% & 91.2\% & 86.3\% & 81.0\% & 75.0\% & 4.29 \\  
& Seer  \cite{tian2024seer}         & $\checkmark$  & Cont. & $\times$ & 96.3\% & 91.6\% & 86.1\% & 80.3\% & 74.0\% & 4.28 \\ 
& \textbf{FLOWER (ours)}  & $\times$   & Flow & $\checkmark$ & \textbf{99.3\%} & \textbf{96.0\%} & \textbf{90.3\%} & \textbf{82.3\%} & \textbf{75.5\%} & \textbf{4.44}$\pm$\textbf{0.04} \\ 
& \textbf{FLOWER (ours)}  & $\checkmark$   & Flow & $\checkmark$ & \textbf{99.4\%} & \textbf{95.8\%} & \textbf{90.7\%} & \textbf{84.9\%} & \textbf{77.8\%} & \textbf{4.53}$\pm$\textbf{0.04} \\ 
\midrule
\multirow{5}{*}{ABCD$\rightarrow$D} & Diff-P-CNN \cite{chi2023diffusionpolicy} & $\times$ & Diffusion & $\times$ & 86.3\% & 72.7\% & 60.1\% & 51.2\% &  41.7\% & 3.16$\pm$0.06 \\
& RoboFlamingo \cite{li2023vision}  & $\times$  & Cont.  & $\checkmark$ & 96.4\% & 89.6\% & 82.4\% & 74.0\% & 66.0\% & 4.09 \\
& DeerVLA \cite{yue2024deer}  & $\times$  & Cont.  & $\checkmark$ & 99.1\% & 93.3\% & 82.1\% & 74.6\% & 63.8\% & 4.13 \\
&GR-1 \cite{wu2023unleashing}        & $\checkmark$  & Cont. & $\times$ & 94.9\% & 89.6\% & 84.4\% & 78.9\% & 73.1\% & 4.21 \\
&MDT   \cite{reuss2024multimodal}   & $\times$      & Diffusion  & $\times$ & 98.6\% & 95.8\% & 91.6\% & 86.2\% & 80.1\%  & 4.52$\pm$0.02 \\  
& MoDE \cite{reuss2024efficient}           & $\checkmark$  & Diffusion & $\times$ & 97.1\% & 92.5\% & 87.9\% & 83.5\% & 77.9\% & 4.39$\pm$0.04 \\
& \textbf{FLOWER (ours)} & $\times$   & Flow & $\checkmark$ & \textbf{98.9\%} & \textbf{96.7\%} & \textbf{93.9\%} & \textbf{90.2\%} & \textbf{85.5\%} & \textbf{4.62}$\pm$\textbf{0.03} \\ 
& \textbf{FLOWER (ours)} & $\checkmark$   & Flow & $\checkmark$ & \textbf{99.2\%} & \textbf{96.9\%} & \textbf{96.9\%} & \textbf{92.3\%} & \textbf{88.3\%} & \textbf{4.67}$\pm$\textbf{0.04} \\ 
\midrule
\multirow{2}{*}{D$\rightarrow$D} &  MDT   \cite{reuss2024multimodal}   & $\times$      & Diffusion  & $\times$ & 93.7\% & 84.5\% & 74.1\% & \textbf{64.4\%} & 55.6\% & 3.72$\pm$(0.05) \\
& RoboUniView \cite{li2024RoboUniView} & $\checkmark$ & Cont  & $\checkmark$ & 96.2\% & 88.8\% & 77.6\% & 66.6\% & 56.3\% & 3.85 \\
& \textbf{FLOWER (ours)}  & $\checkmark$   & Flow & $\checkmark$ & \textbf{97.4\%} & \textbf{92.4\%} & \textbf{86.9\%} & \textbf{81.3\%} & \textbf{74.9\%} & \textbf{4.35}$\pm$\textbf{0.02} \\ 
\bottomrule
\end{tabular}
}
\caption{
\textbf{CALVIN Benchmark results for D, ABC and ABCD.}
The table reports average success rates for individual tasks within instruction chains and the average rollout length (Avg. Len.) to complete 5 consecutive instructions, based on 1000 chains.
Zero standard deviation indicates methods without reported standard deviations.
}
\label{tab: CALVIN Long horizon}
\end{table*}

\textbf{CALVIN Benchmark.} \cite{mees2022calvin}
A language-conditioned manipulation benchmark containing 24k human-teleoperated demonstrations. Each trajectory spans up to 64 timesteps and encompasses 34 predefined basic skills, including tasks such as "rotate blue block right," "move slider right," "lift red block slider," "turn off light bulb," "open drawer," and "place in drawer." The dataset is divided into four splits (A, B, C, D) and evaluates agents on completing 5 consecutive tasks. For evaluation, an agent must complete a sequence of 5 randomly sampled tasks in order (e.g., "open drawer" → "lift blue block drawer" → "place in drawer" → "close drawer" → "turn on light bulb"). The evaluation consists of 1000 rollouts on split D, measuring both the success rate of completing the entire sequence and the average number of successfully completed tasks within each sequence. The Franka Emika Panda robot is controlled via Delta-End-Effector Space with a discrete gripper, utilizing both static and wrist cameras for scene understanding.
There are three benchmark types: D$\rightarrow$D, ABC$\rightarrow$D, ABCD$\rightarrow$D, that depend on the used dataset for trainign the policies. After training all are evaluated on the same environment D. 

For all experiments settings, we train \gls{flore} for up to 40k steps across 4 GPUS with a batch size fo 8 each.
The standardized evaluation protocol enable us to directly compare the results of \gls{flore} against other baselines, which enables a fair comparison to prove the SoTA performance of \gls{flore}.

\textbf{Baselines}
We compare \gls{flore} against a diverse set of available baselines on all CALVIN variants. 
Relevant \gls{vla}s include RoboDual \cite{bu2024towards}, OpenVLA \cite{kim2024openvla}, RoboFlamingo \cite{li2023vision} and depth-reconstruction pretrained RoboUniView \cite{li2024RoboUniView}.
In addition we consider video-based policies pretrained on diverse video data SeeR \cite{tian2024seer}, Video-Prediction-Policy \cite{hu2024videopredictionpolicygeneralist} and GR-1 \cite{wu2023unleashing}.
Moreover, we consider relevant diffusion-based policies like MoDE \cite{reuss2024efficient}, MDT \cite{reuss2024multimodal} and depth-based 3D-Diffusor-Actor \cite{ke2024d} and SuSIE \cite{black2023zero}.
\gls{flore} surpasses all of these baselines across all CALVIN variants with remarkable efficiency in just 6 hours of finetuning.

\subsection{SIMPLER Benchmark Tasks}
\label{sec:app-simpler}
The real2sim benchmark SIMPLER \cite{li24simpler} consists of two evaluation challenges, that we describe in detail below:

\textbf{Google Robot Setting.}
The Google Robot setting comprises four distinct manipulation tasks of varying complexity. The first task, ``pick coke can,'' requires the robot to grasp and lift an empty Coke can from the table. This task includes 75 total trials, testing three different can orientations: horizontally laying, vertically laying, and standing upright. For each orientation, the can is placed at 25 specific grid points within a defined rectangular area on the table, with the environment kept free of distracting elements in its standard configuration.

The second task, ``move objects near objects,'' evaluates the robot's ability to perform relative object positioning with 60 total trials. The setup involves three objects arranged in a triangle pattern, where one object serves as the source, another as the target, and the third as a distractor. The task utilizes eight distinct objects: blue plastic bottle, Pepsi can, orange, 7up can, apple, sponge, Coke can, and Redbull can. Five random triplets are selected from this object pool, with each triplet tested in both upright and inverted triangle patterns. The specific triplet combinations include: (1) blue plastic bottle, Pepsi can, and orange; (2) 7up can, apple, and sponge; (3) Coke can, Redbull can, and apple; (4) sponge, blue plastic bottle, and 7up can; and (5) orange, Pepsi can, and Redbull can.

The third task focuses on drawer manipulation, comprising 54 trials that test the robot's ability to handle articulated objects. The robot is positioned at nine different locations within a rectangular area on the floor and must either open or close a specific drawer (top, middle, or bottom) of a cabinet. This creates a comprehensive evaluation across different robot positions and drawer configurations.

The fourth task combines drawer manipulation with object placement in a 27-trial multi-step interaction. The robot must first open the top drawer and then transfer an apple from the cabinet surface into the drawer. This task evaluates the robot's capability to execute sequential actions, with the robot positioned at three distinct locations and the apple placed at nine specific grid points on the cabinet surface.

\textbf{WidowX + Bridge Setting.}
The WidowX + Bridge setting features four manipulation tasks, each designed to test different aspects of robotic control. The first task, ``put spoon on towel,'' requires placing a spoon from one corner to another of a 15cm square on the tabletop. The spoon's initial orientation alternates between horizontal and vertical, necessitating appropriate gripper reorientation. This task comprises 24 trials total.

The second task, ``put carrot on plate,'' follows a similar structure to the spoon task but replaces the objects, using a carrot instead of a spoon and a plate instead of a towel. This variation tests the robot's ability to transfer manipulation skills to different objects while maintaining the same spatial constraints.

The third task evaluates precise object stacking, requiring the robot to place a green block (3cm in size) on top of a yellow block. The task includes two square configurations with 10cm and 20cm side lengths, creating different spatial challenges. The blocks are positioned at different corners of these squares, totaling 24 trials.

The final task is ``put eggplant into yellow basket,''. The eggplant is randomly positioned within the right basin of a sink, while a yellow basket is placed in the left basin. The eggplant's placement varies in both location and orientation but is carefully arranged to remain easily graspable, avoiding proximity to the sink's edges. This task also comprises 24 trials.

\begin{table}[ht]
    \centering
    \scalebox{0.7}{
    \begin{tabular}{l|cccc|c}
       Method   & Put Carrot on Plate & Spoon on Towel & Stack the Blocks & Eggplant in Yellow Basket  & Average \\
       \midrule
        RT-1-X  & 4 & 0 & 0 & 0 & 1.1 \\
        Octo   & 8 & 12 & 0 & 43 & 16 \\
        CrossFormer  &  15 & 15 & 0 & \textbf{92} & 30 \\
        OpenVLA   & 0 & 0  & 0 & 4  & 1.0 \\
        \textbf{FLOWER}   & 13  &  \textbf{71}  & \textbf{8} & 88  &  \textbf{45} \\
        \bottomrule
    \end{tabular}
    }
    \caption{\textbf{Experimental Results for the SIMPLER Bridge Benchmark.} Average Performance comparison across all Tasks in the Bridge Setting.}
    \label{tab:simpler_bridge}
\end{table}

\begin{table}[ht]
    \centering
    \scalebox{0.7}{
    \begin{tabular}{l|cccc|c}
       Method   & Open/Close Drawer  & Move Near & Open Top Drawer and Place Apple & Pick Coke Can  & Average \\
       \midrule
        RT-1-X  & \textbf{59.7} & 31.7 & \textbf{21.3} & \textbf{56.7} & \textbf{42.4} \\
        Octo  & 22.7 &  4.2  & 0.0 & 17.0 & 11.0 \\
        CrossFormer  & 0.5 & 4.6 & 0.0 & 0.0 & 1.3 \\
        OpenVLA   & 35.6 & \textbf{46.2} & 0.0 & 16.3 & 24.5 \\
        \textbf{FLOWER Cross-X Pret}  & 27.8  & 43.3 & 0.0  &  \textbf{56.3}  & 31.9 \\
        \bottomrule
    \end{tabular}
    }
    \caption{\textbf{Experimental Results for the SIMPLER Google Robot Benchmark.} Average Performance comparison across different task variations for the Google Robot Setting. All tasks have been tested for Visual Matching and Visual Aggregations Variants and show the average performance across both.}
    \label{tab:simpler_google}
\end{table}

We evaluate \gls{flore} on the SIMPLER benchmark \cite{li24simpler} after pretraining on our cross-embodiment mix. 
SIMPLER is a real2sim benchmark that implements several scenes from the diverse BridgeV2 \cite{walke2023bridgedata} and Google Robot setup \cite{brohan2023rt} to test foundation policies after pretraining.
The benchmark requires policies to run approximately $3000$ rollouts in different settings across two benchmarks with 8 different tasks in various conditions.

\textbf{Baselines.} On this setup, we compare \gls{flore} against RT-1X \cite{open_x_embodiment_rt_x_2023}, Octo \cite{octo_2023}, OpenVLA \cite{kim2024openvla}, and CrossFormer \cite{doshiscaling}. For each model, we test on the full benchmark and report the average results for a fair comparison.

\textbf{Results.} The results for the Google Robot tasks are summarized in \autoref{tab:simpler_google} and the results for the Bridge challenge are shown in \autoref{tab:simpler_bridge}. 
Overall, \gls{flore} outperforms both Octo and OpenVLA on both benchmarks, despite having only $200$ GPU hours of pre-training on heterogeneous robot datasets. 
Notably, the \gls{flore} variant achieves stronger overall performance, with gains on the Bridge Benchmark. 
In contrast, on the Google Robot benchmark, RT-1X attains the highest performance across several tasks, suggesting that further improvements in action space modeling or pretraining diversity might be beneficial.
However, \gls{flore} achieves the second best performance and surpasses all other generalist policies in this setting.

These findings show that \gls{flore} delivers strong performance with low computational demands \textbf{(RQ I)} as well as robustly handles diverse robot embodiments and action spaces \textbf{(RQ II)}. The robust performance on the Bridge Benchmark highlights its capability to manage diverse action spaces after heterogeneous pretraining, whereas the areas of lower performance on the Google Robot benchmark point to opportunities for further refinement.

\subsubsection{LIBERO Benchmark.}  
\label{subsec:libero}
The LIBERO benchmark \cite{liu2024libero} comprises multiple task suites testing different aspects of robotic manipulation. LIBERO-10 provides 50 demonstrations for 10 tasks, while LIBERO-90 extends to 90 different tasks. Both versions use a Franka Emika Panda robot with end-effector control and dual camera inputs (static and wrist). The benchmark includes five distinct suites: Spatial (testing spatial relationships), Goal (varying objectives), Object (object manipulation), Long (extended task duration), and Suite-90 (diverse short-horizon tasks). 
Each task is evaluated over 50 trials for each with different starting positions.
We finetune FLOWER with 30k training steps on each setting and 50k steps for LIBERO 90 given its increased size. 

We compare \gls{flore} against both generalist policies such as $\pi_0$ \cite{black2024pi_0} and $\pi_0$-FAST \cite{pertsch2025fast}, the improved $\pi_{0.5}$-ki (knowledge insulation) \cite{driess2025knowledge} OpenVLA \cite{kim2024openvla}, OpenVLA-OFT \cite{kim2025fine} Chain-of-Affordance-VLA (CoA-VLA) \cite{li2024improving}, Octo \cite{octo_2023}, MiniVLA as well as against the current state-of-the-art specialist policy Baku \cite{haldar2024baku}, which uses a small transformer-based model with action chunking. 
As shown in \autoref{tab:libero_results}, \gls{flore} significantly outperforms all baselines across every LIBERO variant, achieving near-perfect completion rates with success rates consistently above $93$\%. 
Notably, on LIBERO-Long, \gls{flore} is the only policy to exceed a $90$\% success rate ($93.5$\%), while other generalist approaches struggle with these complex, long-horizon tasks ($50$-$54$\% success rates), with only the specialist Baku model achieving competitive performance in this demanding setting.

\begin{table*}[ht]
\centering
\scalebox{0.65}{
\begin{tabular}{l|c|c|c|c|c|c}
& Spatial & Object & Goal & Long & 90 & Average (without 90) \\
& SR ($\uparrow$) & SR ($\uparrow$) & SR ($\uparrow$) & SR ($\uparrow$) & SR ($\uparrow$) & SR ($\uparrow$) \\
\hline
Diff-P-CNN \cite{chi2023diffusionpolicy} & 78.3 $\pm$ 1.1\% & 92.5 $\pm$ 0.7\% & 68.3 $\pm$ 1.2\% & 50.5 $\pm$ 1.3\% & - & 72.4 $\pm$ 0.7\% \\
Octo \cite{octo_2023} & 78.9 $\pm$ 1.0\% & 85.7 $\pm$ 0.9\% & 84.6 $\pm$ 0.9\% & 51.1 $\pm$ 1.3\% & - & 75.1 $\pm$ 0.6\% \\
OpenVLA \cite{kim2024openvla}  & 84.7 $\pm$ 0.9\% & 88.4 $\pm$ 0.8\% & 79.2 $\pm$ 1.0\% & 53.7 $\pm$ 1.3\% & - & 76.5 $\pm$ 0.6\% \\
OpenVLA-OFT \cite{kim2025fine} & 97.6\% & 98.4\% & \textbf{97.9\%} & 94.5\% & - & 97.1\% \\
CoA-VLA \cite{li2024improving} & 85.3 $\pm$ 0.9\% & 93.1 $\pm$ 1.0\% & 85.8 $\pm$ 0.9\% & 55.0 $\pm$ 1.2\% & - & 79.8 $\pm$ 0.5\% \\
Baku \cite{haldar2024baku} & - & - & - & 86.0\% & 90.0\% & - \\
MiniVLA  & - & - & - & - & 86.0\% & - \\
MoDE \cite{reuss2024efficient} & - & - & - & 94.0\% & 95.0\% & - \\
$\pi_0$ \cite{black2024pi_0} & 96.8\% & 98.8\% & 95.8\% & 85.2\% & - & 94.2\% \\
$\pi_0$-FAST \cite{pertsch2025fast} & 96.4\% & 96.8\% & 88.6\% & 60.2\% & - & 85.5\% \\
$\pi_{0.5}$-ki (from scratch) \cite{driess2025knowledge} & 96.6\% & 97.2\% & 94.6\% & 84.8\% & 92.7\% & 93.3\% \\
$\pi_{0.5}$-ki (from generalist model) \cite{driess2025knowledge} & \textbf{98.0\%} & 97.8\% & 95.6\% & 85.8\% & \textbf{96.0\%} & 94.3\% \\
\textbf{FLOWER} & 97.5 $\pm$ 0.8\% & \textbf{99.1 $\pm$ 0.4\%} & 96.1 $\pm$ 0.6\% & \textbf{94.9 $\pm$ 1.2\%} & 94.7 $\pm$ 1.0\% & \textbf{96.9 $\pm$ 0.7\%} \\
\hline
\end{tabular}}
\caption{\textbf{Experimental Results for the LIBERO Benchmarks.} SR: Success Rate. Best results in each column are shown in bold. FLOWER achieves state-of-the-art results across most tested settings.}
\label{tab:libero_results}
\end{table*}

\subsubsection{Aloha Benchmark.} 
\label{subsec:aloha}
The Aloha benchmark \cite{zhao2023learning} provides 50 human-collected demonstrations for two tasks: Cube Transfer and Insertion. In both tasks, a bi-manual Aloha robot operating in joint space is equipped with a single top-view camera and proprioceptive state input. We evaluate each task over 500 episodes, with each episode consisting of 500 steps. For the baselines we adopt ACT\cite{zhao2023learning} and Diffusion Policy\cite{chi2023diffusionpolicy} implemented by lerobot \cite{cadene2024lerobot}.

\begin{figure}
    \centering
    \includegraphics[width=\linewidth]{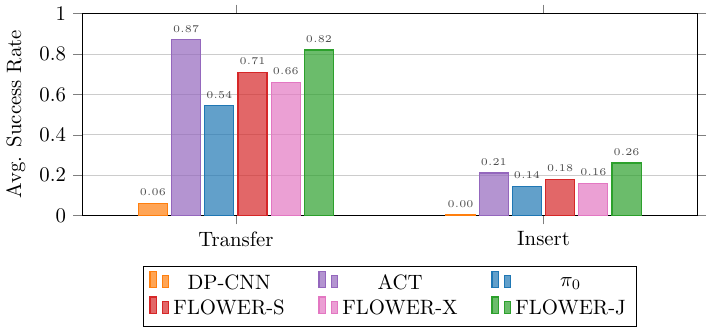}
    \caption{\textbf{\emph{Aloha Simulation Tasks:}} Average success rates of different models over 500 evaluations on \textit{Cube Transfer} and \textit{Insertion}. \emph{S} denotes FLOWER trained from scratch, \emph{X} applies cross-action space pretraining, and \emph{J} represents droid-only joint state pretraining.}
    \label{fig:aloha_sim_tasks}
\end{figure}

Next, we evaluate \gls{flore}'s ability to learn challenging high frequency control on the Aloha simulation setup \cite{zhao2023learning}. 
We test several versions of \gls{flore} that have been finetuned on different pretraining mixes: Cross-X and joint state droid only. 
In addition, we compare \gls{flore} against two common specialist policies: Diffusion Policies and the state-of-the-art policy for bi-manual setups Action Chunking Tranformer (ACT) \cite{zhao2023learning}.
We use the two simulation tasks, "Insert the peg into the socket." and "Pick up the cube with the right arm and transfer it to the left arm.", that are visualized in \autoref{fig:environments}.
The dataset contains $50$ human-collected demonstrations for each tasks.

\textbf{Results.}
As shown in \autoref{fig:aloha_sim_tasks}, \gls{flore} achieves a strong performance on both tasks with eight denoising steps and outperforms the specialist ACT policy on the challenging Insertion task by a considerable margin. 
\gls{flore} achieves comparable performance on the Transfer task compared to ACT expect for the variant pretrained on cross action space data. 
The standard Diffusion Policy is not able to solve any of the tasks. 
Comparing the different pretraining versions of \gls{flore}, we find that the joint only mix using droid achieves the best results by a considerable margin.
Surprisingly, the cross-embodied pretraining is not able to achieve strong results and its final performance is even lower than the version trained from scratch. 

\subsection{Real Kitchen Play Dataset.} 
\label{sec:app-kitchen-play}
We conducted data collection through teleoperation, utilizing a leader-follower robot configuration to ensure precision and intuitive control \citep{jiang2024comprehensive}. The dataset includes proprioceptive sensor readings and images captured by two static cameras at a frequency of 6 Hz. Actions were represented as normalized desired joint positions.
In total, we curated 417 labeled short-horizon segments, each paired with text instructions. To enhance diversity in task descriptions, GPT-4 was employed to generate varied language annotations.

\textbf{Evaluation Protocol.}
Each policy is tested $5$ times for each task from a starting position not seen in training with some added noise to it.
During our experiments, we further varied the orientation of the banana slightly for the robot to pick up, while we kept the toaster in the same position during all our experiments.
We report the average success rate and rank for each task to determine the best policy.

\subsubsection{Pretraining Details for Baselines for the Real World Kitchen}
\label{subsec:finetuning-baselines-real-world}
For finetuning the baseline generalist robot policies (Octo, CrossFormer, and OpenVLA), we adhered as closely as possible to the official recommendations provided in their respective GitHub repositories, with only minimal modifications where necessary. 
All experiments were conducted on a single-node GPU cluster equipped with four NVIDIA RTX 4090 GPUs (24GB each).

For \textbf{Octo}, we fine-tuned the model for 50,000 steps, which took approximately 6 hours. In our setup, we used two images per sample—designating the top image as \texttt{image\_primary} and the side image as \texttt{image\_secondary}. 
The baseline Octo model has a default action space of 7 dimensions (delta end-effector position, rotation, and gripper controls). To accommodate our tasks, we extended the default action head by one additional dimension, creating a 7+1 dimensional absolute action space before fine-tuning.

For \textbf{CrossFormer}, we again relied on the default fine-tuning settings from the original repository and trained for 50,000 steps, which took around 12 hours. 
The image setup was identical to that used for Octo. We introduced a new action head, \texttt{new\_arm\_single\_joint}, with an action dimension of 8, and developed a new observation tokenizer specifically for the secondary image. All other modules were initialized from the pretrained weights provided in the repository and then fine-tuned.

For \textbf{OpenVLA}, which originally supports only delta end-effector actions and enforces an assertion to prevent the use of joint-space OXE datasets for pretraining or fine-tuning, we modified the code to remove this assertion. We then introduced appropriate action and normalization masks (masking nothing for the action and masking the gripper for normalization). Using the default fine-tuning configuration with LoRA-based updates, we trained OpenVLA for 150,000 steps. 
Due to memory constraints, we reduced the batch size from the default of 16 to 1 and applied gradient accumulation over 4 steps (as opposed to a larger accumulation factor, which would have substantially increased training time). 
This fine-tuning process took approximately 60 hours on our 4-GPU setup. 

For \gls{flore} we finetuned our model on the kitchen dataset for 50,000 steps. Since \gls{flore} has been pretrained on single image, we extended the second static image for finetuning. No additional modifications have been made to guarantee a fair comparison against the baselines. 

Overall, these modifications allowed us to fine-tune all baseline models under comparable conditions (approximately 100k–150k steps, moderate batch sizes, and consistent GPU resources), ensuring a fair evaluation on our real-world kitchen tasks.

\subsubsection{Failure Cases for different Policies}

\textbf{Octo.}
The most common failure mode involves Octo fixating on the microwave - repeatedly opening, closing, or attempting to interact with its door even when the task involves other objects or locations. 
The second most frequent failure involves Octo's poor object manipulation, particularly with the pot and banana, where it either drops items prematurely or fails to lift them high enough to clear obstacles like the sink edge. 
Finally, there's a consistent pattern of spatial navigation issues where Octo either pushes objects into walls, hovers aimlessly above target locations, or places objects in incorrect intermediate positions (like between the sink and stove).

\textbf{CrossFormer.}
The Crossformer policy exhibits several consistent failure patterns across tasks, including freezing in place, hovering without executing actions, and getting stuck on objects (e.g., sink, microwave door, oven). 
Many failures involve misinterpreting tasks, such as repeatedly pretending to place toast in the sink or confusing objects like the banana and the oven tray. The model also struggles with manipulating objects correctly, often failing to grasp, dropping, or pushing objects off surfaces rather than placing them accurately. Additionally, it frequently interacts with unintended objects, such as opening and closing the microwave

\textbf{OpenVLA.}
OpenVLA frequently fails due to object manipulation errors, such as pushing, flipping, or throwing objects off surfaces rather than placing them correctly. 
A recurring issue is poor grasping ability, especially with pots and bananas, often failing to lift them or dropping them prematurely. 
Additionally, the policy exhibits random movement behaviors, such as hovering aimlessly, crashing into the kitchen, or moving without executing the task. It also struggles with partial execution, frequently opening and then immediately closing doors or trays instead of completing the full action.

\textbf{\gls{flore}.}
The most common failure mode of \gls{flore} is imprecise spatial positioning, particularly evident in tasks like pushing the toaster lever where the agent consistently misses by about 1cm. We hypothesize that this is due to workspace normalization issues at boundary regions. 
The second major failure pattern involves interaction with pots in the sink, where FLOWER either gets stuck in loops just before completion, fails to properly clear the sink walls, or incorrectly routes objects (like trying to drop pots into the sink during stove-to-stove transfers). 
Finally, there are issues with excessive force application in some cases, particularly with the toaster where the agent occasionally rips it off rather than interacting with it properly.

\subsection{Generalization Experiments}
\label{subsec:generalization-tests}
Finally, we evaluate \gls{flore} against the best baseline in several generalization experiments. In these experiments, we test the models under conditions that introduce variations not encountered during training. Table~\ref{tab:generalization_comparison} reports the performance of the different methods on tasks such as “Move Pot from Right Stove to Sink”, “Open Oven”, and “Pull Oven Tray” under three scenarios: Novel Object, Flashlight, and Background Distractors. For instance, in the Novel Object condition, new object instances are introduced, while the Flashlight and Background Distractors settings simulate changes in illumination and environmental clutter. These settings collectively challenge the models to generalize beyond their training distribution.

\begin{figure*}[ht]
    \centering
    \begin{subfigure}{0.45\linewidth}
        \centering
        \includegraphics[width=0.9\linewidth]{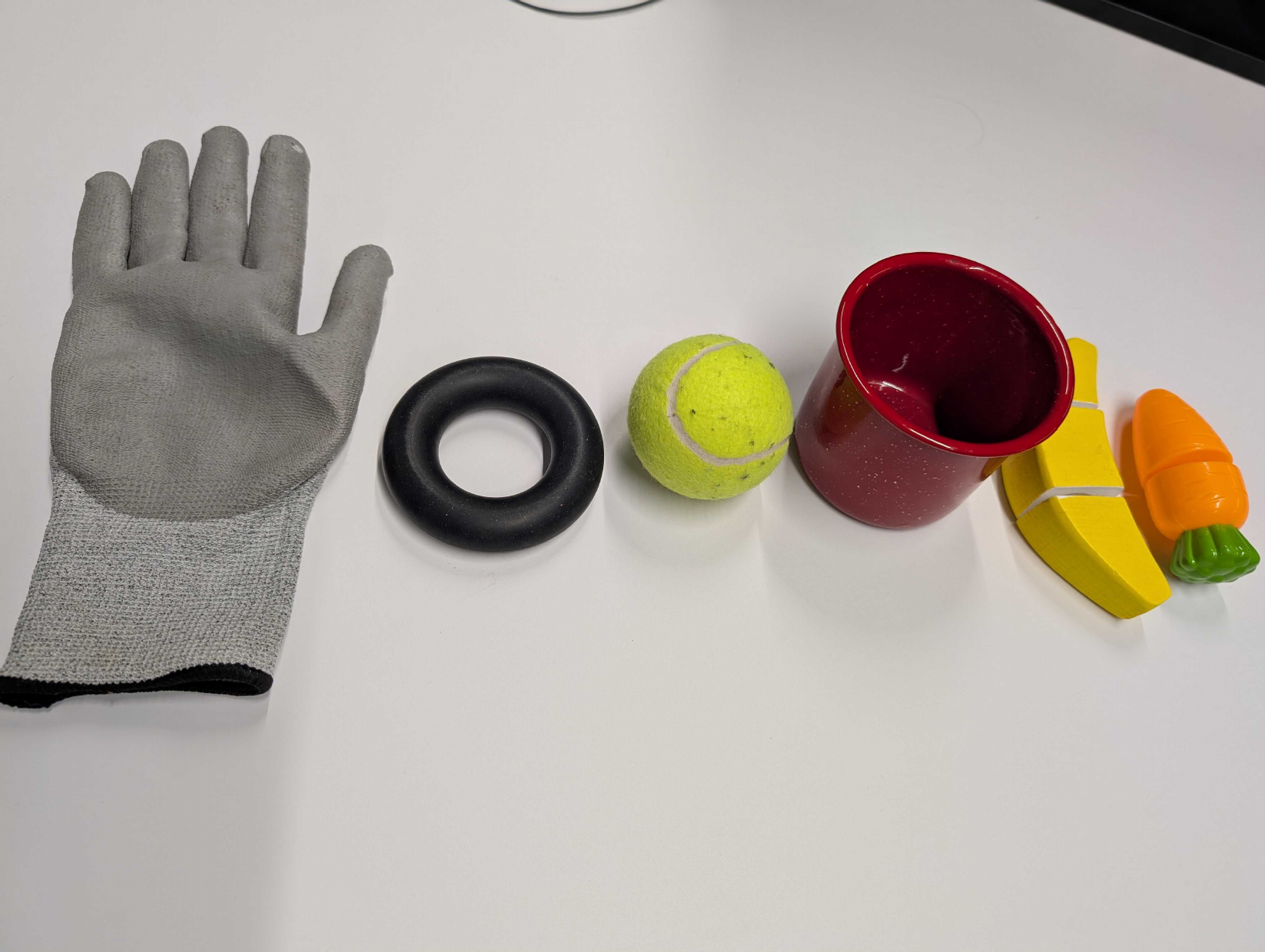}
        \caption{Unseen Objects for Generalization Experiments}
        \label{fig:novel-objects}
    \end{subfigure}
    \hfill
    \begin{subfigure}{0.45\linewidth}
        \centering
        \includegraphics[width=0.9\linewidth]{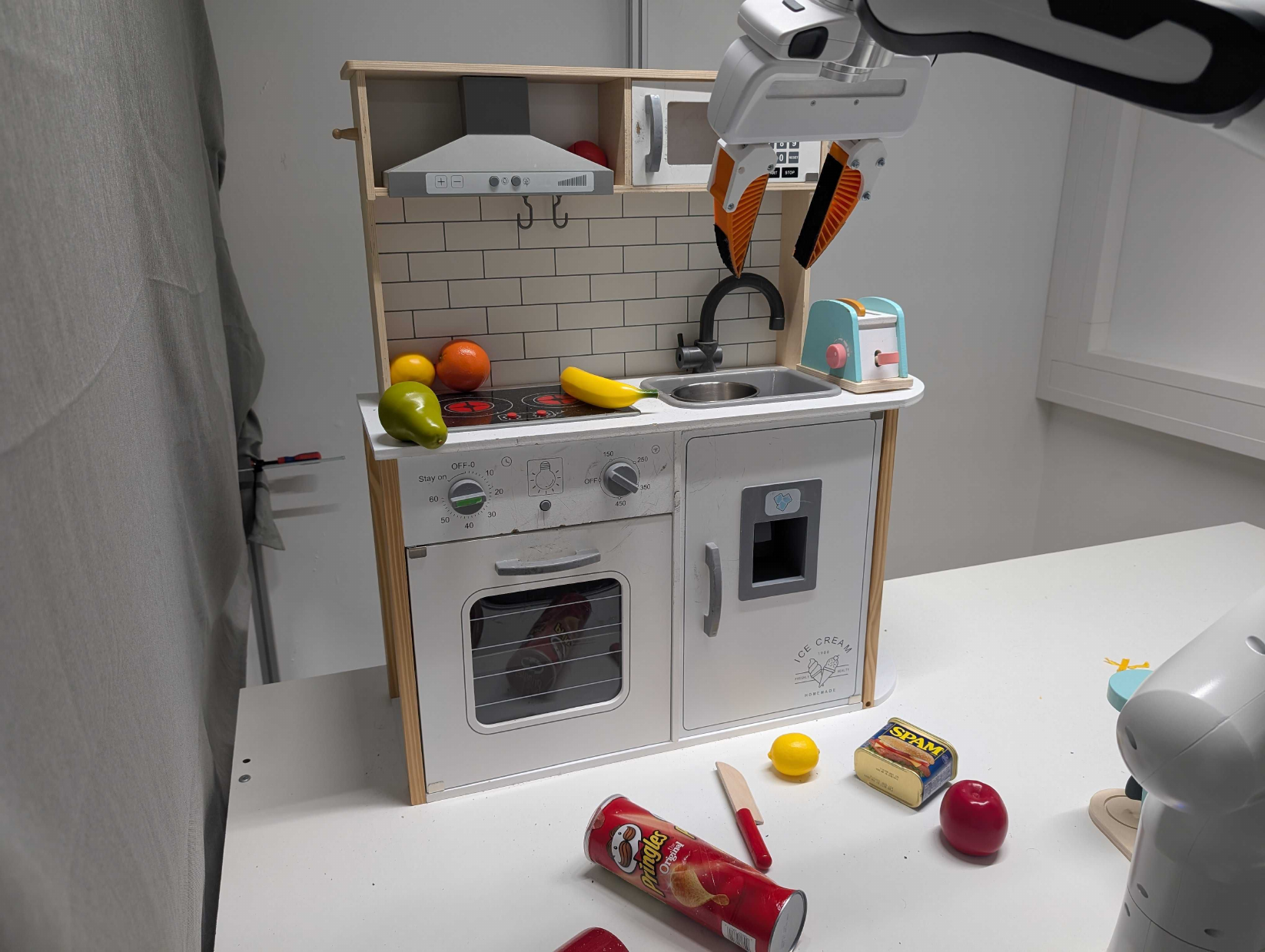}
        \caption{Cluttered Scene with Distractors}
        \label{fig:cluttered-scene}
    \end{subfigure}
    \caption{\textbf{Generalization Experiments:} Examples of unseen objects (left) and cluttered scenes (right) used to test the adaptability of the policies in our real-world setting. All the tested objects are not included in the training dataset.}
    \label{fig:combined-generalization-test}
\end{figure*}

In particular, our experiments also examine the models' abilities to manipulate novel objects—those not present in the initial training distribution. 
The objects that are new to the model are highlighted in \textbf{bold} in Table~\ref{tab:generalization_comparison}. As shown, FLOWER consistently achieves higher success rates and lower ranks when manipulating these unfamiliar items, demonstrating robust performance even under significant distribution shifts. \autoref{fig:combined-generalization-test} provides visual examples of these novel objects and the scene with various background distractions. 
This additional analysis underscores the strength of our approach in adapting to unseen variations in both object appearance and environmental context.

\subsubsection{Novel Task Compositions}
\label{subsec:compostional-experiments}
To further evaluate our method’s capacity for compositional generalization, we designed a set of novel task compositions that require the agent to combine multiple subtasks into a coherent, long-horizon plan. Each task is defined as a sequence of actions that must be executed in a specific order. For instance, the \textbf{Sequence: Open and Close All Appliances} task comprises the following subtasks: “Open the Microwave”, “Open the Oven”, “Open the Ice”, “Close the Ice”, “Close the Oven”, and “Close the Microwave”. This sequence challenges the model to manipulate various kitchen appliances in a coordinated manner, ensuring that the prescribed order of operations is maintained under varying conditions.

In addition, we introduced two other sequence tasks to test different aspects of compositionality. The \textbf{Sequence: Move Items Between Stovetop and Sink} task requires the agent to transfer items between workstations, with subtasks including “Move Banana from Right Stove to Sink”, “Push the Toaster Lever”, “Move Pot from Left Stove to Right Stove”, “Pick Up Toast and Place it at the Sink”, “Move Pot from Right Stove to Left Stove”, and “Move Banana from Sink to Right Stove”. Finally, the \textbf{Sequence: Operate the Oven} task focuses on oven manipulation and is composed of the subtasks “Open the Oven”, “Pull the Oven Tray”, “Move Banana from Right Stove to Oven Tray”, “Push the Oven Tray”, and “Close the Oven”. These novel task compositions simulate realistic, multi-step scenarios and provide a rigorous benchmark for evaluating the ability of our approach to integrate learned sub-skills into coherent, long-horizon behaviors.

\begin{table*}
\centering
\begin{tabular}{l|c|c|c|c|c}
\toprule
\multirow{2}{*}{\textbf{Task}} & \multirow{2}{*}{\textbf{SR/R}} & \textbf{Octo} & \textbf{OpenVLA} & \textbf{CrossFormer} & \textbf{FLOWER} \\
\midrule
\multirow{2}{*}{Pot from right stove to sink} & SR & 0.0\% & 60.0\% & \textbf{100.0}\% & 80.0\% \\
& R & 4 & 3 & \textbf{1} & 2 \\
\multirow{2}{*}{Pot from sink to right stove} & SR & 0.0\% & 0.0\% & 0.0\% & \textbf{20.0}\% \\
& R & 2 & 2 & 2 & \textbf{1} \\
\multirow{2}{*}{Open oven} & SR & 40.0\% & 0.0\% & 0.0\% & \textbf{60.0}\% \\
& R & 2 & 3 & 3 & \textbf{1} \\
\multirow{2}{*}{Pull oven tray} & SR & 0.0\% & 40.0\% & 0.0\% & \textbf{100.0}\% \\
& R & 3 & 2 & 3 & \textbf{1} \\
\multirow{2}{*}{Open microwave} & SR & 40.0\% & \textbf{100.0}\% & 40.0\% & \textbf{100.0}\% \\
& R & 3 & \textbf{1} & 3 & \textbf{1} \\
\multirow{2}{*}{Close microwave} & SR & 20.0\% & 80.0\% & 40.0\% & \textbf{100.0}\% \\
& R & 4 & 2 & 3 & \textbf{1} \\
\multirow{2}{*}{Banana from right stove to sink} & SR & 10.0\% & 40.0\% & 40.0\% & \textbf{100.0}\% \\
& R & 4 & 2 & 2 & \textbf{1} \\
\multirow{2}{*}{Banana from sink to right stove} & SR & 0.0\% & 0.0\% & 20.0\% & \textbf{60.0}\% \\
& R & 3 & 3 & 2 & \textbf{1} \\
\multirow{2}{*}{Push toaster lever} & SR & 0.0\% & \textbf{100.0}\% & 0.0\% & 0.0\% \\
& R & 2 & \textbf{1} & 2 & 2 \\
\multirow{2}{*}{Pickup toast and put to sink} & SR & 0.0\% & 20.0\% & \textbf{80.0}\% & 40.0\% \\
& R & 4 & 3 & \textbf{1} & 2 \\
\multirow{2}{*}{Open Ice} & SR & 0.0\% & 0.0\% & 0.0\% & \textbf{100.0}\% \\
& R & 2 & 2 & 2 & \textbf{1} \\
\multirow{2}{*}{Banana from right stove to oven tray} & SR & 0.0\% & 0.0\% & 0.0\% & \textbf{40.0}\% \\
& R & 2 & 2 & 2 & \textbf{1} \\
\multirow{2}{*}{Pot from sink to left stove} & SR & \textbf{0.0}\% & \textbf{0.0}\% & \textbf{0.0}\% & \textbf{0.0}\% \\
& R & \textbf{1} & \textbf{1} & \textbf{1} & \textbf{1} \\
\multirow{2}{*}{Pot from left stove to right stove} & SR & 40.0\% & 0.0\% & 20.0\% & \textbf{80.0}\% \\
& R & 2 & 4 & 3 & \textbf{1} \\
\multirow{2}{*}{Banana from tray to right stove} & SR & \textbf{0.0}\% & \textbf{0.0}\% & \textbf{0.0}\% & \textbf{0.0}\% \\
& R & \textbf{1} & \textbf{1} & \textbf{1} & \textbf{1} \\
\multirow{2}{*}{Close oven} & SR & 40.0\% & \textbf{100.0}\% & 20.0\% & 80.0\% \\
& R & 3 & \textbf{1} & 4 & 2 \\
\multirow{2}{*}{Pot from right stove to left stove} & SR & 30.0\% & 0.0\% & 0.0\% & \textbf{40.0}\% \\
& R & 2 & 3 & 3 & \textbf{1} \\
\multirow{2}{*}{Push oven tray} & SR & 0.0\% & 20.0\% & \textbf{40.0}\% & 20.0\% \\
& R & 4 & 2 & \textbf{1} & 2 \\
\multirow{2}{*}{Pot from left stove to sink} & SR & 0.0\% & 80.0\% & 40.0\% & \textbf{100.0}\% \\
& R & 4 & 2 & 3 & \textbf{1} \\
\multirow{2}{*}{Close Ice} & SR & 0.0\% & 0.0\% & 0.0\% & \textbf{100.0}\% \\
& R & 2.0 & 2 & 2 & \textbf{1} \\
\midrule
\multirow{2}{*}{\textbf{Overall Performance}} & SR & 10\% & 31\% & 22\% & \textbf{61\%} \\
& R & 2.70 & 2.10 & 2.20 & \textbf{1.25} \\
\bottomrule
\end{tabular}
\caption{\textbf{Detailed Results for all tested Real Robot Tasks in the Kitchen Environment.} Each task has two rows: the first (SR) reports success rate (\%), and the second (R) reports rank within that task (lower rank = better performance). The best results per task are highlighted in \textbf{bold}.}
\label{tab:updated_success_rank}
\end{table*}

\begin{table*}[h]
    \centering
    \resizebox{\textwidth}{!}{%
    \begin{tabular}{l
                    cc  
                    cc  
                    cc  
                    }
        \toprule
        \multirow{2}{*}{\textbf{Task}} & \multicolumn{2}{c}{\textbf{Novel Object}} & \multicolumn{2}{c}{\textbf{Flashlight}} & \multicolumn{2}{c}{\textbf{BG Distractors}} \\
        \cmidrule(lr){2-3} \cmidrule(lr){4-5} \cmidrule(lr){6-7}
         & \textbf{FLOWER} & \textbf{OpenVLA} & \textbf{FLOWER} & \textbf{OpenVLA} & \textbf{FLOWER} & \textbf{OpenVLA} \\
        \midrule
        Move the \textbf{black donut} from sink to right stove & \textbf{33.3}  & 0.0   & -     & -     & -     & -     \\
        Move the \textbf{tennis ball} from right stove to sink  & \textbf{66.7}  & 33.3  & -     & -     & -     & -     \\
        Move the \textbf{tennis ball} from sink to right stove   & 0.0   & 0.0   & -     & -     & -     & -     \\
        Move the \textbf{black donut} from right stove to sink   & \textbf{33.3}  & 0.0   & -     & -     & -     & -     \\
        Move the \textbf{red cup} from right stove to sink       & 33.3  & \textbf{66.7}  & -     & -     & -     & -     \\
        Move the \textbf{glove} from sink to right stove         & \textbf{33.3}  & 0.0   & -     & -     & -     & -     \\
        Move the \textbf{carrot} from right stove to sink        & \textbf{33.3}  & 0.0   & -     & -     & -     & -     \\
        Move the \textbf{glove} from right stove to sink         & \textbf{100}   & 0.0   & -     & -     & -     & -     \\
        Move the \textbf{carrot} from sink to right stove        & 0.0   & 0.0   & -     & -     & -     & -     \\
        Move the \textbf{red cup} from sink to right stove        & 0.0   & 0.0   & -     & -     & -     & -     \\
        \midrule
        Open the microwave       & -   & -   & \textbf{100}   & \textbf{100}   & \textbf{100}   & \textbf{100}   \\
        Pull the oven tray                            & -   & -   & \textbf{100}   & 0.0   & \textbf{100}   & 66.7  \\
        Move banana from right stove to sink              & -   & -  & \textbf{100}   & 33.3  & 66.7  & 66.7  \\
        Close the oven                                    & -  & -  & 33.3  & \textbf{66.7}  & 66.7  & 0.0   \\
        Push down the toaster lever                           & -   & -   & 0.0   & 0.0   & 33.3  & \textbf{100}   \\
        Move pot from right stove to sink                & -  & -  & -  & 0.0   & \textbf{66.7}  & 33.3  \\
        Open the ice box                                       & -   & -   & 0.0   & 0.0   & 66.7  & 0.0   \\
        Open the oven                                      & -   & -   & \textbf{100}   & 0.0   & \textbf{100}   & 0.0   \\
        Close the microwave                            & -   & -   & \textbf{100}   & \textbf{100}   & \textbf{100}   & 66.7  \\
        Move pot from left stove to sink                 & -   & -   & 0.0   & 0.0   & 33.3  & 66.7  \\
        Close the ice box                                & -   & -   & 0.0   & 0.0   & \textbf{100}   & 0.0   \\
        Pick up toast and put it in the sink              & -   & -   & 0.0   & 0.0   & 0.0   & 0.0   \\
        \midrule
        \textbf{Average} 
         & \textbf{33.3} & 10.0  
         & \textbf{50.0} & 25.0  
         & \textbf{69.5} & 41.7  
         \\
        \bottomrule
    \end{tabular}%
    }
    \caption{\textbf{Generalization experimental results for novel objects, distractions and new lighting conditions.} The table reports the success rate (in \%) of the corresponding policy evaluated under three different generalization scenarios: Novel Object, Flashlight, and Background Distractors (evaluated 3 times for each setting). The best score for each test is highlighted in \textbf{bold}. A dash (\texttt{-}) indicates that the task was not evaluated in that scenario.}
    \label{tab:generalization_comparison}
\end{table*}

\begin{table*}[h]
\centering
\scalebox{0.8}{
\begin{tabular}{ll|ccccc|c}
\toprule
\textbf{Task} & \textbf{Method} & \textbf{1} & \textbf{2} & \textbf{3} & \textbf{4} & \textbf{5} & \textbf{Avg. Seq. Len.} \\  
\midrule
\multirow{2}{*}{Seq: Stovetop + Sink} 
    & FLOWER   & \textbf{66.7\%}  & \textbf{66.7\%}  & \textbf{66.7\%}  & \textbf{33.3\%}  & \textbf{33.3\%}  & \textbf{2.67} \\
    & OpenVLA  & \textbf{66.7\%}  & 33.3\%  & 33.3\%  & 0.0\%   & --      & 1.33 \\
\midrule
\multirow{2}{*}{Seq: Open Close All} 
    & FLOWER   & \textbf{100\%}   & \textbf{100\%}   & \textbf{100\%}   & \textbf{100\%}   & \textbf{100\%}   & \textbf{5.00} \\
    & OpenVLA  & 33.3\%  & 0.0\%   & --      & --      & --      & 0.33 \\
\midrule
\multirow{2}{*}{Seq: Oven} 
    & FLOWER   & 0.0\%   & --      & --      & --      & --      & 0.00 \\
    & OpenVLA  & 0.0\%   & --      & --      & --      & --      & 0.00 \\
\midrule
\textbf{Overall Performance (FLOWER)} &  & \textbf{51.1\%} & --- & --- & --- & --- & \textbf{2.56} \\
\textbf{Overall Performance (OpenVLA)} &  & \textbf{16.7\%} & --- & --- & --- & --- & \textbf{0.55} \\
\bottomrule
\end{tabular}
}
\caption{\textbf{Long Horizon Task Composition Results.} For each sequence task, the per-instruction success rates (in \%) are shown for the first 5 instructions (if applicable) along with the average sequence length. “--” indicates that no instruction was successfully solved at that index. The Overall Performance rows report the average success rate (computed over all available instructions) and the average sequence length across tasks for each method.}
\label{tab:long_horizon_sequences}
\end{table*}

\begin{figure*}
    \centering
    \includegraphics[width=0.92\linewidth]{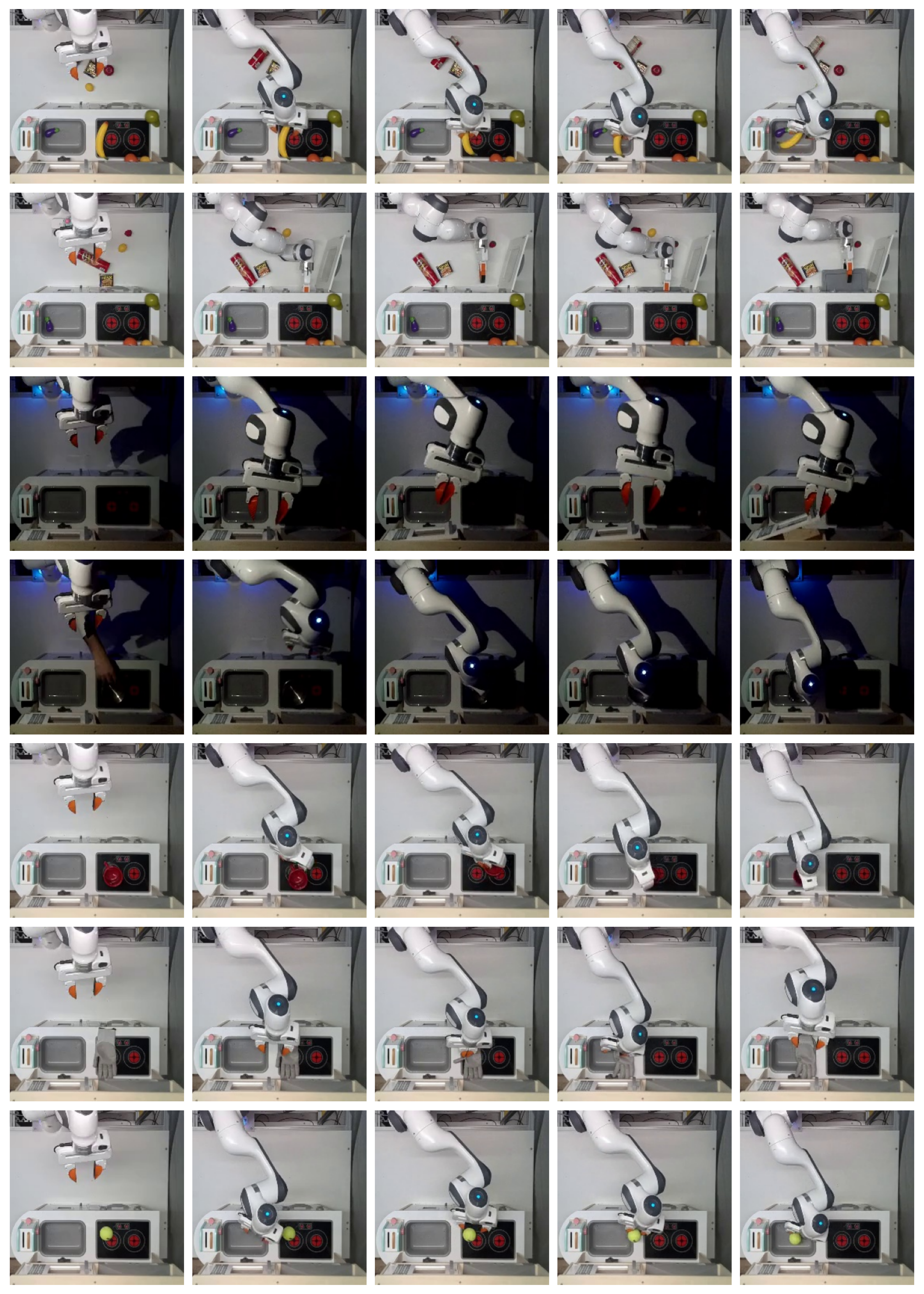}
    \caption{\textbf{Example generalization rollouts.} First two rows show rollouts with background distractors, rows 3 and 4 show rollouts with only a flashlight as a light source, and the last 3 rows showcase novel objects.}
    \label{fig:rollout-figure-generalization}
\end{figure*}

\section{Extended Related Work}

\textbf{Cross-Embodiment Learning}
A core challenge in robotics is learning a unified policy for heterogeneous embodiments with distinct action and sensor spaces. Early approaches often applied modular policies~\cite{huang2020one} or hardware-conditioned representations~\cite{chen2018hardware}, and some leveraged graph-based representations to generalize across different robot hands~\cite{patel2024get, yang2023polybot}. However, these efforts tended to focus on smaller datasets or simplified environments.
Recent work on large-scale cross-embodiment includes RoboCat~\cite{bousmalis2023robocat} and PolyBoT~\cite{yang2023polybot}, which use action tokenization or hierarchical controllers, respectively. 
\citet{liu2024rdt} propose a unified 258-dimensional action space (RDT-1B) with fixed action prediction length of $64$ for all action spaces, while CrossFormer~\cite{doshiscaling} employs separate action heads with a continuous action prediction head for different embodiments but lacks a pretrained vision-language component for generalization to diverse instructions. 
By incorporating an action-type Global AdaLN conditioned Flow Transformer with a pretrained VLM, \gls{flore} efficiently handles multiple embodiments while maintaining both action expressiveness and semantic understanding.

\textbf{Rectified Flow and Diffusion Models in Robotics.}
Diffusion models~\cite{ho2020denoising, song2020score, song2021denoising} have become widely used for generating continuous robot actions from visual inputs~\cite{chi2023diffusionpolicy, octo_2023, reuss2023goal, ke20243d}, offering multi-modal behavior~\cite{jia2024towards} and good scaling with large datasets~\cite{octo_2023, liu2024rdt, reuss2024efficient}. 
More recently, \emph{Rectified Flow}~\cite{esser2024scaling, albergo2022building, lipman2022flow} has emerged as a promising alternative, enabling a straight-line probability path for action sampling that requires few discretization steps.
In robotic policy learning, ActionFlow~\cite{funk2024actionflow},  $\pi_0$ \cite{black2024pi_0} and others~\cite{zhang2024affordance, braun2024riemannian} showed rectified flow can generate actions more rapidly than standard diffusion. 
Such fast inference is crucial for high-frequency robot setups like Aloha \cite{fu2024mobile}. 
Yet, these works are typically confined to a single embodiment or a single action space. 
By contrast, \gls{flore} is the first completely open-source policy to apply rectified flow as a \emph{generalist} policy component, unifying expressive and  multimodal flow-based action generation with diverse vision-language contexts. While $\pi_0$ published their weights for finetuning, the code for pretraining and their dataset remains closed source.



\end{document}